\documentclass[11pt]{article}

\usepackage{pdfpages}
\usepackage{amsmath, amsfonts, amssymb, bm, url}
\usepackage{graphicx}
\usepackage[a4paper, total={7.75in, 9.25in}]{geometry}
\usepackage[utf8]{inputenc}
 \usepackage[T1]{fontenc}
\usepackage{authblk}

\usepackage{hyperref}
\usepackage{multicol}
\newcommand{\sfig}[1]{Fig.~S#1}
\newcommand{\stab}[1]{Table~S#1}
\newcommand{\smat}{Suppl. material}

\begin{document}

\title{\emph{lgpr:} An interpretable nonparametric method for inferring \\covariate effects from longitudinal data}

\author[1]{Juho Timonen}
\author[1]{Henrik Mannerstr\"om}
\author[1]{ Aki Vehtari}
\author[1]{Harri L\"ahdesm\"aki}
\affil[1]{Department of Computer Science, Aalto University}

\maketitle

\abstract{
Longitudinal study designs are indispensable for studying disease progression. Inferring covariate effects from longitudinal data, however, requires interpretable methods that can model complicated covariance structures and detect nonlinear effects of both categorical and continuous covariates, as well as their interactions. Detecting disease effects is hindered by the fact that they often occur rapidly near the disease initiation time, and this time point cannot be exactly observed. An additional challenge is that the effect magnitude can be heterogeneous over the subjects. We present \textit{lgpr}, a widely applicable and interpretable method for nonparametric analysis of longitudinal data using additive Gaussian processes. We demonstrate that it outperforms previous approaches in identifying the relevant categorical and continuous covariates in various settings. Furthermore, it implements important novel features, including the ability to account for the heterogeneity of covariate effects, their temporal uncertainty,  and appropriate observation models for different types of biomedical data. The \textit{lgpr} tool is implemented as a comprehensive and user-friendly R-package. \textit{lgpr} is available at \url{jtimonen.github.io/lgpr-usage} with documentation, tutorials, test data, and code for reproducing the experiments of this paper.
}

\vspace{0.5cm}

\begin{multicols}{2}

\section{Introduction}
\label{s: introduction}

Biomedical studies often collect observational longitudinal data, where the same individuals are measured at several time points. This is an important study design for examining disease development and has been extensively leveraged in biomedical studies, including various -omics studies, such as proteomics \cite{liu2018}, metagenomics \cite{vatanen2016}, and single-cell transcriptomics \cite{sharma2018}. The measured response variable of interest can be continuous (such as the abundance of a protein), discrete (such as the number of sequencing reads in a genomic region), or binary (such as patient condition). Often also several additional variables -- i.e. covariates -- are measured for each subject at each measurement time point. These can be categorical variables (such as sex, location, or whether the subject is diagnosed with a disease or not) or continuous (such as age, time from disease initiation, or blood pressure). Identifying the relevant covariates that affect the response variable is important for assessing potential risk factors of the disease and for understanding disease pathogenesis.

A large body of literature has focused on the statistical analysis of longitudinal data \cite{diggle2002}. Observations corresponding to the same individual are intercorrelated, and specialized statistical methods are therefore required. Methods must be able to model both time-dependent and static covariate effects at the same time and handle irregular measurement intervals, missing data, and a varying number of measurements for different individuals. Generalized linear mixed (GLM) models \cite{stroup2012} have been found to best conform to these challenges, and they have become the standard workhorse for longitudinal data analysis. 
 
 The R-package \textit{lme4} \cite{bates2015} has gained high popularity and become a default choice for fitting GLM models. These models, however, require specifying a parametric (linear) form for the covariate effects and provide biased inferences when their true effects are nonlinear or nonstationary.

A Gaussian process (GP) is a popular Bayesian nonparametric model that is commonly used for time series modeling \cite{rasmussen2006, roberts2013}. Properties of a GP are determined by the used kernel function, and with multivariate input data, GPs can be made more interpretable by defining an additive kernel function \cite{duvenaud2011}. Additive GP methods that are particularly designed for longitudinal study designs include the method in \cite{quintana2016} and \textit{LonGP} \cite{cheng2019}. The latter one is specifically designed for model selection and uses a forward search algorithm that adds new additive components one by one. Due to computational convenience, GP models such as \textit{LonGP} are often restricted to Gaussian observation model, which is not appropriate for count or proportion data commonly observed in biomedicine. A common approach is to use the Gaussian observation model after first applying a variance-stabilizing transform, such as log-transform, to the response variable, but this is not statistically justified and can lead to biased inferences \cite{ohara2010}. See \smat \ for more background information and related research.

Longitudinal studies often comprise a case and control group, and commonly a clinically determined time of disease initiation for each case individual is marked in the data. To reveal phenomena related to disease progression or to identify biomarkers, statistical modeling can utilize the disease-related age, i.e.\ time from disease initiation or onset, as one covariate that can explain changes in the response variable. Disease effects can be rapid when compared to other effects and expected to occur near the time of disease initiation, which is another aspect that GLM models cannot capture. A major challenge is that many diseases, such as Type 1 Diabetes (T1D), are heterogeneous \cite{pietropaolo2007}, and disease-specific biomarkers are likely
detectable only in a subset of the diagnosed individuals. Another problem that can confound the analysis of disease effects, is that the disease initiation (or onset) time is difficult to determine exactly. For example in Type 1 Diabetes (T1D), the presence of islet cell autoantibodies in the blood is the earliest known marker of disease initiation \cite{ziegler2013}, but they can only be measured when the subject visits a doctor. In general, the detected disease initiation time can differ from the true initiation time, and the extent of this difference can vary across individuals. To our knowledge, there exist no methods that can model nonstationary disease effects while taking into account the disease heterogeneity and uncertainty of initiation time.

\begin{figure}% Fig. 1
\centering
\includegraphics[width=0.9\textwidth]{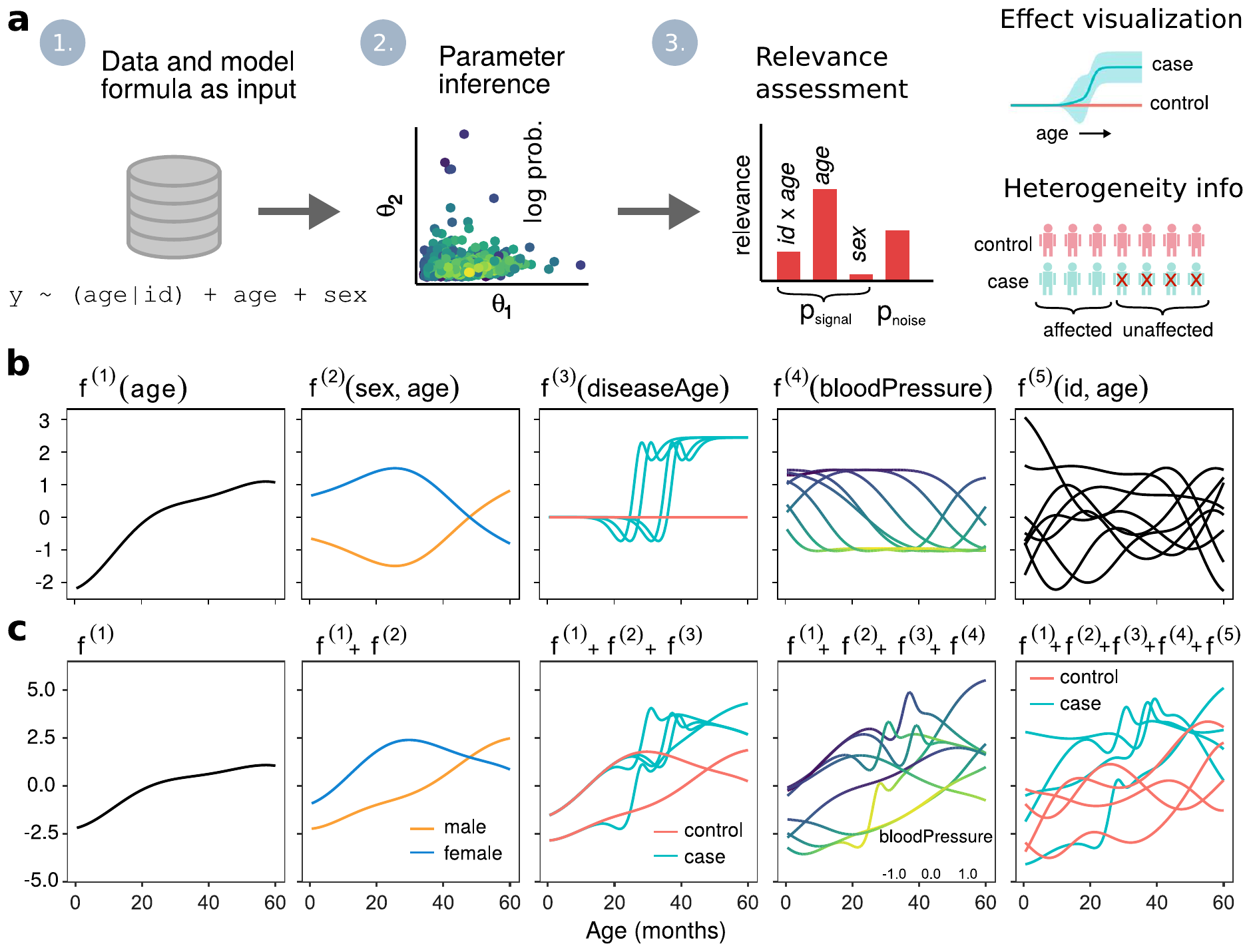}
\caption{\label{fig: main1} Overview of additive Gaussian process modeling of longitudinal data using \textit{lgpr}. \textbf{a)} A typical workflow with \textit{lgpr}. \textit{1.} User gives the data and model formula as input, along with possible additional modeling options such as non-default parameter priors or a discrete observation model. \textit{2.} The model is fitted by sampling the posterior distribution of its parameters. \textit{3.} Relevances of different covariates and interaction terms are computed. The inferred signal components can be visualized to study the magnitude and temporal aspects of different covariate effects. If a heterogeneous model component was specified, the results inform how strongly each individual experiences the effect. \textbf{b)} Examples of different types of covariate effects that can be modeled using \textit{lgpr}. The components $f^{(j)}$, $j=1, \ldots, 5$ are draws from different Gaussian process priors. This artificial data comprises 8 individuals (4 male, 4 female), and 2 individuals of each sex are cases. The shown age-dependent components are a shared age effect $f^{(1)}$, a sex-specific deviation $f^{(2)}$ from the shared age effect, a disease-related age ($diseaseAge$) effect $f^{(3)}$, and a subject-specific deviation $f^{(5)}$ from the shared age effect. For each of the diseased individuals, the disease initiation occurs at a slightly different age, between 20 and 40 months. Here, the magnitude of the disease effect is equal for each case individual, but \textit{lgpr} can model also heterogeneous effects. The component $f^{(4)}$ is a function of blood pressure only, but is plotted against age for consistency as the simulated blood pressure variable has a temporal trend. \textbf{c)} The cumulative effect $f = \sum_j f^{(j)}$ is a sum of the low-dimensional components.}
\end{figure}

In this work, we propose a longitudinal data analysis method called \textit{lgpr}, designed for revealing general nonlinear and nonstationary effects of individual covariates and their interactions (see Fig.~\ref{fig: main1}a). It is based on the additive GP approach similar to  \textit{LonGP} but provides several significant improvements that tackle the challenges stated above. We use special interaction kernels that allow separating category effects (e.g. different temporal profiles for male and female subjects) from shared effects. This allows us to develop a straightforward but useful covariate relevance assessment method, which requires fitting only one model and gives estimates of the proportion of variance explained by each signal component and noise. Our package implements additive GP modeling and covariate relevance assessment also in the case of a non-Gaussian observation model and allows incorporating sample normalization factors that account for technical effects commonly present for example in RNA-sequencing data. Additionally, our tool can account for uncertainty in the disease effect time and features a novel kernel that allows identification of heterogeneous effects detectable only in a subset of individuals.  For increased interpretability of disease effects, we propose a new variance masking kernel which separates effects related to disease development from the baseline difference between case and control individuals.

We have implemented \textit{lgpr} as a user-friendly R-package \cite{rcore2018} that can be used as a plug-in replacement for \textit{lme4}. Under the hood, Bayesian model inference is carried out using the dynamic Hamiltonian Monte Carlo sampler \cite{hoffman2014, betancourt2017}, as implemented in the high-performance statistical computation framework Stan \cite{carpenter2017}. The new tool is summarized in Fig.~\ref{fig: main1}a.

We use simulated data to prove the benefit of each new feature of our method. Additionally, we use \textit{lgpr} to analyse data from two recent T1D studies. The first one is a longitudinal proteomics data set \cite{liu2018} and the second one is RNA-sequencing data from peripheral blood cells \cite{kallionpaa2019}.

\section{Methods}
\label{s: methods}

\subsection{The probabilistic model}
We denote a longitudinal data set with $N$ data points and $D$ covariates % and one response variable
by a tuple $(\bm{X}, \bm{y})$, where $\bm{X}$ is an ${N \times D}$ covariate matrix and  $\bm{y} \in \mathbb{R}^N$ contains the response variable measurements. We refer to the $i$th row of $\bm{X}$ by $\bm{x}_i \in \mathcal{X}$, where $\mathcal{X} = \times_{d=1}^D \mathcal{X}_d$ and $\mathcal{X}_d$ is the set of possible values for covariate $d$. In general, $\mathcal{X}_d$ can be discrete, such as the set of individual identifiers, or connected such as $\mathbb{R}$ for (normalized) age.
%The number of measurements can differ over the subjects, and the measurement times can have irregular intervals. 
Our model involves  an unobserved signal $f: \mathcal{X} \rightarrow \mathbb{R}$, which is a function of the covariates. The signal is linked to $\bm{y}$ through a likelihood function, motivated by a statistical observation model for $\bm{y}$, and uses transformed signal values $g^{-1}\left(  f(\bm{x}_i)+ c_i \right)$, where $g$ is a link function and $c_i$ are possible additional scaling factors. We have implemented inference under Gaussian, Poisson,  binomial, beta binomial (BB) and negative binomial (NB) likelihoods, and they are defined in detail in \smat.
%  $f(\bm{x}) \sim \mathcal{GP}\left(m(\bm{x}), k(\bm{x}, \bm{x}') \right)$ 

The process $f$ is assumed to consist of $J$ low-dimensional additive components, so that $f(\bm{x}) = f^{(1)}(\bm{x}) + \ldots + f^{(J)}(\bm{x})$  (see Fig.~\ref{fig: main1}b-c). Each component $j$ is modeled as Gaussian process (GP) with zero mean function and kernel function $\alpha_j^2 k_j (\bm{x}, \bm{x}')$. This means that the vector of function values $\bm{f}^{(j)} = \left[f^{(j)} (\bm{x}_1), \ldots, f^{(j)} (\bm{x}_N) \right]^\top$ has a multivariate normal prior $\bm{f}^{(j)}  \sim \mathcal{N}\left(\bm{0},\bm{K}^{(j)} \right)$
with zero mean vector and $N \times N$ covariance matrix with entries $\{ \bm{K}^{(j)} \}_{ik} = \alpha_j^2 k_j(\bm{x}_i, \bm{x}_k)$.  Because the components are \textit{a priori} independent, the sum $f$ is also a zero-mean Gaussian process with kernel $k(\bm{x},\bm{x}')  = \sum_{j=1}^J \alpha_j^2 k_j(\bm{x},\bm{x}’)$. See more info about GPs in \smat  \ or \cite{rasmussen2006}. 

The parameter $\alpha_j^2$ is called the marginal variance of component $f^{(j)}$ and it determines how largely the component varies. The base kernel function $k_j(\bm{x},\bm{x}’)$ on the other hand determines the component's shape, as well as covariance structure over individuals or groups  (see Fig.~\ref{fig: main1}b-c).  The base kernels are constructed, as explained in the next section, so that each $f^{(j)}, j = 1, \ldots, J$ is a function of only one or two covariates. This is a sensible assumption in many real-world applications and apt to learn long-range structures in the data \cite{duvenaud2011}. Furthermore, this decomposition into additive components allows us to obtain interpretable covariate effects after fitting the model. Duvenaud \textit{et al.} (2011) used also higher-order interaction terms (which we could incorporate into our model as well), but they did not study relevances of individuals covariates, as high-order interactions inherently confound their interpretation.

\subsection{Kernel functions for longitudinal data}

\subsubsection{Shared effects}
Stationary shared effects of continuous covariates are modeled using the exponentiated quadratic kernel $k_{\text{eq}}(x,x' \mid \ell) = \exp \left( -\frac{(x-x')^2}{2 \ell^2}\right)$ . Here, $x$ refers to a generic continuous covariate, and each shared effect component has its own lengthscale parameter $\ell$, which determines how rapidly the component can vary. For example, a shared age effect kernel is $ k_{\text{eq}}(x_{\text{age}},x_{\text{age}}' \mid \ell_{\text{age}})$.

\subsubsection{Category effects}
%We assume that for any given individual, the value of a categorical covariate $z$ is constant over all observations. 
Effects of categorical covariates (such as sex or individual id) can be modeled either as fluctuating category-specific deviations from a shared effect (interaction of a categorical and continuous covariate) or as static category-specific offsets. For a pair of categorical covariate $z$ (with $M\geq2$ categories) and continuous covariate $x$, we use the kernel function
\begin{equation}
   \label{eq: gp_anova}
    k_{z \times x}( (z,x), (z',x')  \mid \ell) =  k_{\text{zerosum}}(z, z') \cdot k_{\text{eq}}(x, x' \mid \ell),
\end{equation}
when modeling the effect of $z$ as deviation from the shared effect of $x$. The zero-sum kernel $k_{\text{zerosum}}(z, z') $, returns $1$ if $z = z'$ and $  \frac{1}{1 - M}$ otherwise. This is similar to the GP ANOVA approach in \cite{kaufman2010}. If $f:\mathbb{R} \times \{1, \ldots, M\} \rightarrow \mathbb{R}$ is modeled using the kernel in Eq.~\ref{eq: gp_anova}, the sum $\sum_{r=1}^M f(t,r)$ is always zero for any $t$ (see proof in \smat). %This means that $f$ represents a category-specific deviation from the shared effect (see \sfig{1} for illustration). 
The fact that the sum over categories equals exactly zero for any $t$ greatly helps model interpretation as this property separates the effect of the categorical covariate from the shared effect (see \sfig{1} for illustration).  If the effect of $z$ is modeled as a batch or group offset, which does not depend on time or other continuous variables, the corresponding kernel function is just $k_{\text{zerosum}}(z, z')$. Again, $z$ refers to a generic categorical covariate.

\subsubsection{Nonstationary effects}
We use the input warping approach \cite{snoek2014} to model  nonstationary functions $f^{(j)}(x)$, where most variability occurs near the event  $x = 0$. The nonstationary kernel is
\begin{equation}
\label{eq: kernel_ns}
    k_{\text{ns}}(x, x' \mid a, \ell) =  k_{\text{eq}}(\omega_a(x), \omega_a(x') \mid \ell), 
\end{equation}
where $\omega: \mathbb{R} \rightarrow ]-1,1[$ is a monotonic nonlinear input warping function
\begin{equation}
    \label{eq: input_warp}
    \omega_a(x) = 2 \cdot \left(\frac{1}{1 + e^{-a x}} - \frac{1}{2} \right),
\end{equation}
and the parameter $a$ controls the width of the effect window around $x = 0$.

\subsubsection{Disease effects}
 In \cite{cheng2019}, disease effects were modeled using the kernel in Eq.~\ref{eq: kernel_ns} for the disease-related age $x_{\text{disAge}}$, i.e. time from disease initiation or onset of each individual. Note that for the control subjects, $x_{\text{disAge}}$ is not observed at all. In general, data for a continuous covariate $x$ can be missing in part of the observations. In such cases, we adopt the approach of \cite{cheng2019} and multiply the kernel of $x$ with a mask kernel which returns $0$ if either of its arguments is missing and $1$ if they are available.

Whereas this approach can model a nonstationary trend that is only present for the diseased individuals, its drawback is that it can capture effects that are merely a different base level between the diseased and healthy individuals. In order to find effects caused by the disease progression, we design a new kernel
\begin{equation}
\label{eq: kernel_varmask}
    k_{\text{vm}}(x, x' \mid a, \ell) = f^a_{\text{vm}}(x) \cdot  f^a_{\text{vm}}(x') \cdot k_{\text{ns}}(x, x' \mid a, \ell), 
\end{equation}
where $f^a_{\text{vm}}(x): \mathbb{R} \rightarrow ]0,1[$ is a variance mask function that forces the disease component to have zero variance, i.e. the same value for both groups, when $x \rightarrow - \infty$. We choose to use $f^a_{\text{vm}}(x) = \frac{1}{1 + e^{-a (x-r)}}$, which means that the allowed amount of variance between these groups rises sigmoidally from 0 to the level determined by the marginal variance parameter, so that the midpoint is at $r = \frac{1}{a} \log \left( \frac{h}{1-h} \right)$ and $\omega(r \mid a) = 2h-1$. The parameter $h$ therefore determines a connection between the regions where the disease component is allowed to vary between the two groups and where it is allowed to vary over time. In our experiments, we use the value $h=0.025$. This means, that 95\% of the variation in $\omega$ occurs on the interval $[-r,r]$. The kernels in Eq.~\ref{eq: kernel_ns} and Eq.~\ref{eq: kernel_varmask} combined with the missing value masking, as well as functions drawn from the corresponding GP priors, are illustrated in \sfig{2}.

\subsubsection{Heterogeneous effects}
To model effects that have the same effect shape but possibly different magnitude for each individual, we define additional parameters $\bm{\beta} = [\beta_1, \ldots, \beta_Q]$, where $Q$ is the number of individuals and each $\beta_i \in [0,1]$. Denote $\mathcal{X}_{\text{id}} = \{1,\ldots,Q\}$ and assume two individuals $x_{\text{id}} = q \in \mathcal{X}_{\text{id}}$ and $x_{\text{id}}' = q' \in \mathcal{X}_{\text{id}}$. An effect is made heterogeneous in magnitude by multiplying its kernel by $k_{\text{heter}}(x_{\text{id}}, x'_{\text{id}} \mid \bm{\beta}) = \sqrt{\beta_q \beta_{q'}}$. For example, to specify a heterogeneous disease effect component, we use the novel kernel
\begin{equation}
\label{eq: kernel_heter_vm}
 k_{\text{heter}}(x_{\text{id}}, x'_{\text{id}} \mid \bm{\beta}) \cdot  k_{\text{vm}}(x_{\text{disAge}}, x'_{\text{disAge}} \mid a, \ell_{\text{disAge}}) .
\end{equation}
For heterogeneous disease effects, the number of needed $\beta$ parameters equals the number of only the case individuals.

In our implementation, the prior for the unknown parameters $\bm{\beta}$ is $\beta_i \sim \text{Beta}(b_1,b_2)$, where the shape parameters $b_1$ and $b_2$ can be defined by the user. By default, we set $b_1 = b_2 = 0.2$, in which case most of the prior mass is near the extremes 0 and 1  (\sfig{3}c). This choice is expected to induce sparsity, so that some individuals have close to zero effect magnitude. The posterior distributions of $\beta_i$ can then be used to make inferences about which case individuals are affected by the disease ($\beta_i$ close to 1) and which are not ($\beta_i$ close to 0). The kernel in Eq.~\ref{eq: kernel_heter_vm} is illustrated in \sfig{2}.

\subsubsection{Temporally uncertain effects}
The presented disease effect modeling approach relies on being able to measure the disease onset or effect time $t_{\text{eff}}$ for each case individual, since the disease-related age is defined as $x_{\text{disAge}} = x_{\text{age}} - t_{\text{eff}}$. In \cite{cheng2019}, $t_{\text{eff}}$ was defined as age on the clinically determined disease initiation date, but in general the effect time can differ from it. Our implementation allows Bayesian inference also for the effect times, and can therefore capture effects that for some or all case individuals occur at a different time point than the clinically determined date. The user can set the prior either directly for the effect times $t_{\text{eff}}$, or for the difference between the effect time and observed initiation time, $\Delta t = t_{\text{obs}} -  t_{\text{eff}}$. The first option is suitable if the disease is known to commence at certain age for all individuals. The latter option is useful in a more realistic setting where such information is not available, and it is reasonable to think that the clinically determined initiation time $t_{\text{obs}}$ is close to the true effect time.

%\section{Algorithm}

\subsection{Model  inference} 
 We collect all marginal variances, lengthscales and other possible kernel hyperparameters in a vector $\bm{\theta}_{\text{kernel}}$. Parameters of the observation model are denoted by $\bm{\theta}_{\text{obs}}$ and other parameters such as those related to input uncertainty by $\bm{\theta}_{\text{other}}$. The collection of all unknown parameters is then $\bm{\theta} = \left \{ \bm{\theta}_{\text{kernel}}, \bm{\theta}_{\text{obs}}, \bm{\theta}_{\text{other}} \right \}$. Under the hood, \textit{lgpr} uses the dynamic Hamiltonian Monte Carlo sampler \cite{hoffman2014} with multinomial sampling of dynamic length trajectories \cite{betancourt2017}, as implemented in Stan \cite{carpenter2017}, to obtain $S$ draws from the posterior distribution of $\bm{\theta}$. The parameters are given robust priors that normalize model fitting (specified in \smat), and our software includes prior predictive checks that help in prior validation. Our default prior for the steepness parameter $a$ of the input warping function (Eq.~\ref{eq: input_warp}) allows disease effects that occur approximately on a 36 month interval around the disease initiation time. \sfig{3}d-e illustrate the effect of the prior choice for this parameter.  

The remaining unknowns of the model are the values of the function components, for which we use the notation $\bm{f}^{(j)} = [f^{(j)}(\bm{x}_1), \ldots, f^{(j)}(\bm{x}_N) ]^{\top},$ where $\bm{x}_i$ is the $i$th row of $\bm{X}$, and $\bm{f} = \sum_{j=1}^J \bm{f}^{(j)}$. Under the Gaussian observation model, the posterior distributions of $\bm{f}^{(1)}, \ldots, \bm{f}^{(J)}$ and $\bm{f}$ can be derived analytically (see \smat). With other observation models, we sample the posterior of each $\bm{f}^{(j)}$ simultaneously with $\bm{\theta}$.

\subsection{Covariate relevance assessment}
Our method only requires sampling the posterior of a full model including all covariates. From now on we assume that each continuous covariate can be present in at most one shared effect term and arbitrarily many interactions terms. Its relevance is then interpreted to be the relevance of the shared effect component. We also assume that each categorical covariate can appear only in one term, which can be an interaction or a first-order term, and its relevance is then interpreted to be the relevance of the component where it appears. This way the covariate relevance assessment problem reduces to determining the relevance of each component.
%covariate relevance assessment procedure relies on building a full model with all available covariates of interest.

After posterior sampling, we have $S$ parameter draws $ \left \{ \bm{\theta}^{(s)} \right\}_{s=1}^{S}$ and if using a non-Gaussian observation model, also draws $\left \{ \bm{f}^{(j, s)} \right \}_{s=1}^S$ of each function component $j=1, \ldots, J$. For each draw $s$, our model gives predictions $\bm{y}^*_s = \left[ y^*_{1,s}, \ldots, y^*_{N,s}\right]$.  With the Gaussian observation model, $\bm{y}^*_s = \bm{\mu}_{s}$, i.e.\ the analytically computed posterior mean.  With other observation models, $\bm{y}^*_s = g^{-1}\left(\bm{h}^{(s)} \right)$, where $\bm{h}^{(s)} = \bm{c} +  \sum_{j=1}^J \bm{f}^{(j, s)}$ and $\bm{c} = \left[c_1,\ldots,c_N\right]$ are the scaling factors.

We determine how much of the data variation is explained by noise,  using an approach closely related to the Bayesian $R^2$-statistic \cite{gelman2019}. The noise proportion in draw $s$ is
\begin{equation} 
\label{eq: p_noise_s}
  p_{\text{noise}}^{(s)} = \frac{RSS_s}{ESS_s + RSS_s} \in [0, 1]
\end{equation}  
where $RSS_s = \sum_{i=1}^N (y^*_{i, s} - y_i)^2$ and $ESS_s = \sum_{i=1}^N (y^*_{i, s} - \bar{y}^*_{s})^2$ are the residual and explained sum of squares, respectively, and $\bar{y}^*_{s} = \frac{1}{N} \sum_{i=1}^N y^*_{i, s} $.  With this definition, $p_{\text{noise}}^{(s)}$ will be one if the model gives constant predictions and zero if predictions match data exactly. Note that with binomial and BB models, $y_i$ is replaced by $y_i/\eta_i$, where $\eta_i$ is the total the number of trials, as $y_i$ is the number of successes. 
 
The proportion of variance that is associated with the actual signal, $p_{\text{signal}}^{(s)}  = 1 - p_{\text{noise}}^{(s)}$, is further divided between each model component. For cleaner notation, we define the variation of a vector $\bm{v} = \left[v_1, \ldots, v_L \right]$ as a sum of squared differences from the mean, i.e. $SS(\bm{v}) = \sum_{l=1}^L \left(v_l - \bar{v_l} \right)^2$. The relevance of component $j$ is 
 \begin{equation}
    \label{eq: relevance_s}
    \text{rel}_j^{(s)} = p_{\text{signal}}^{(s)} \frac{SS_j^{(s)}}{\sum_{j'=1}^J SS_{j'}^{(s)}}
\end{equation}
where  $SS_j^{(s)} = SS \left( \bm{\mu}^{(j, s)} \right)$ with Gaussian observation model and $SS_j^{(s)} = SS \left( \bm{f}^{(j, s)} \right)$ otherwise. Above we used $ \bm{\mu}^{(j, s)}$ to denote the posterior mean vector of component $j$, corresponding to draw $s$ (see \smat). The final component and noise relevances are then
\begin{equation}
    \label{eq: rel}
    \text{rel}_j =  \frac{1}{S} \sum_{s=1}^S \text{rel}_j^{(s)} \hspace{0.5cm} \text{and} \hspace{0.5cm} p_{\text{noise}} =  \frac{1}{S} \sum_{s=1}^S p_{\text{noise}}^{(s)},
\end{equation}
i.e. averages over the $S$ MCMC draws. Our definition has the properties that $\text{rel}_j \in [0,1]$ for all $j$, and that we can compute the proportion of variance explained by a subset of components $\mathcal{J} \subseteq \{1, \ldots, J\}$ simply as $\text{rel}_{\mathcal{J}} = \sum_{j \in \mathcal{J}} \text{rel}_j$. Furthermore, adding more components will always increase the proportion of explained variance. 

Our main focus is on providing these numeric values that describe how much effect each covariate has on the response variable. However, we also provide a method for perfoming covariate selection. The approach is to select the minimal subset of components $\mathcal{J}_{\text{sel}}$ that together with noise explain at least $T \%$ of variance. Formally, $\mathcal{J}_{\text{sel}} = \arg \min_{\mathcal{J}} |\text{rel}_{\mathcal{J}}|$, subject to $\text{rel}_{\mathcal{J}} + p_{\text{noise}} \geq \frac{T}{100}$ and $T=95$ by default.  In \smat, we describe also a probabilistic extension of this method. %The user can change the desired threshold $T$ according their preferences, taking into account the total number of model components and amount of noise in the data.
We emphasize that when selecting covariates, we are not testing whether or not a given effect is exactly zero. Therefore we do not perform multiple testing corrections, as in frequentist literature, when analysing multiple response variables (several proteins or genes). See \cite{gelman2012} for discussion.

A related method, which also relies on selecting a minimal subset of covariates based on inference of a full model with all covariates, is the projection predictive model selection method \cite{goutis1998}. It has been shown to perform well in predictive covariate selection for generalized linear models \cite{piironen2017a}. However, it still requires comparing lots of alternative sub-models to the full model, whereas in our case finding the minimal subset of predictors does not require additional sampling or parameter fitting. %It reduces to just sorting the list of relevances $\{\text{rel}_1, \ldots, \text{rel}_J \}$. 
 Moreover, sequential subset search methods, such as the projection predictive method and \textit{LonGP} \cite{cheng2019}, are prone to most often selecting the most expressive components. We argue that our method is more suitable for longitudinal GP models that contain components of different complexities. For example, an individual-specific age component is more expressive than a shared age effect component.

\section{Results}
\label{s: results}

\subsection{Experiments with simulated data}
\label{s: sim_data}
First, we use simulated data to demonstrate the accuracy of covariate relevance assessment and benefits of the novel features of our method. In each experiment, we generate data with different types of continuous and categorical covariates (see \smat \ for details of data simulation). In order to test the accuracy of our covariate relevance assessment,  we simulate noisy measurements of a response variable so that only part of the covariates are relevant. In each experiment we generate several random data set realizations and measure performance in classifying covariates as relevant or irrelevant using the area under curve (AUC) measure for receiver operating characteristic (ROC) curves. Higher AUC value indicates better performance. The computed covariate relevances (rel$_j$ in Eq. \ref{eq: rel}) are used as a score in the ROC analyses, which are performed using the \textit{pROC} package \cite{robin2011}.

\subsubsection{Comparison with linear mixed effect modeling and LonGP}

\begin{figure} % Fig. 2
\centering
\includegraphics[width=0.8\textwidth]{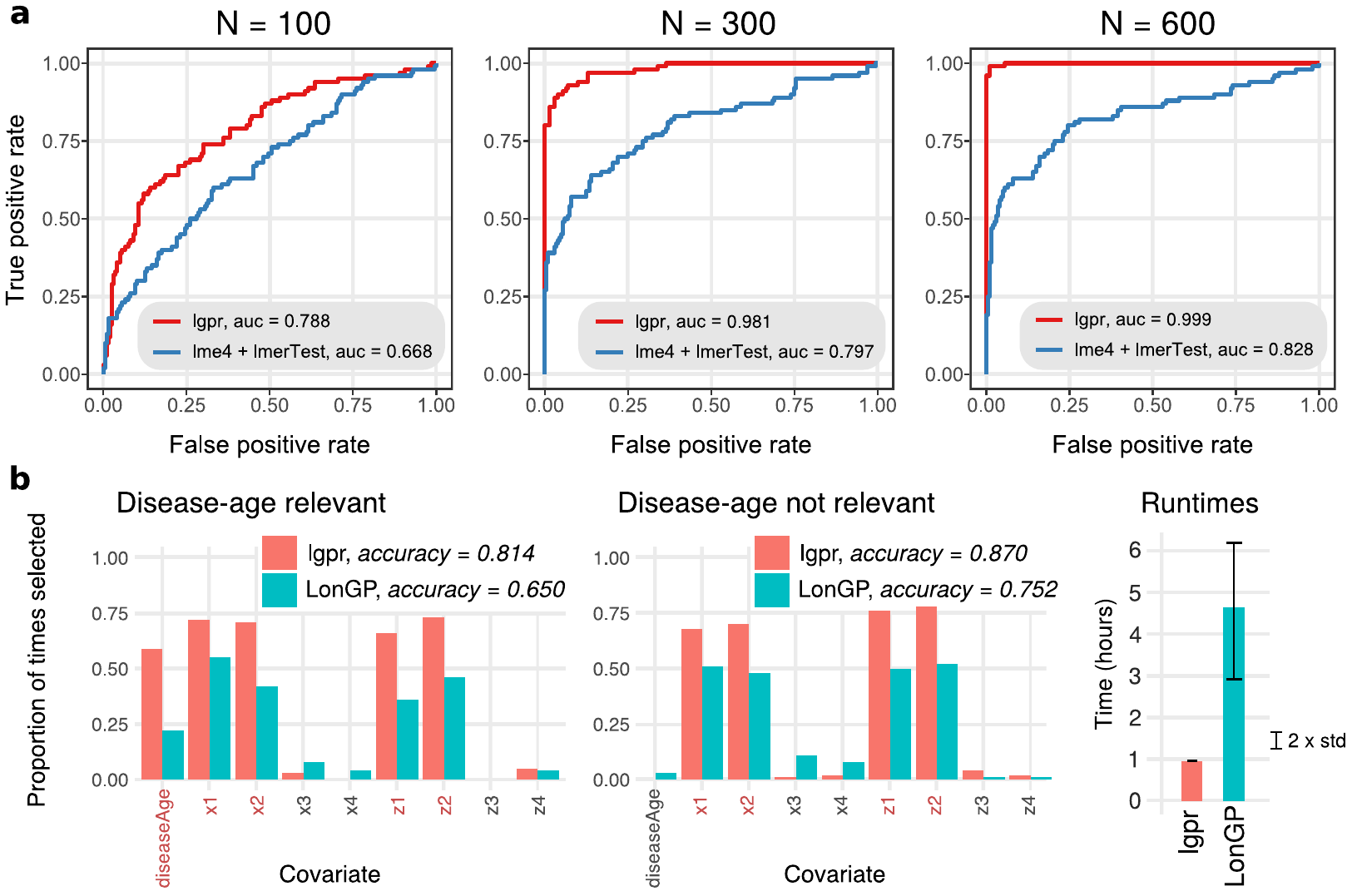}
\caption{\label{fig: comparisons} Covariate relevance assessment comparison with other methods and demonstration of our method's scalability. \textbf{a)} Comparison between \textit{lgpr} and linear mixed effect modeling using the \textit{lme4} and \textit{lmerTest} packages. The panels show ROC curves for the problem of classifying covariates as relevant or irrelevant, when the total number of data points is $N = 100$, $300$ and $600$, respectively. \textbf{b)} Comparison against \textit{LonGP}. The bar plots show the fraction of times each covariate was chosen in the final model over 100 simulated data sets. Red text indicates the covariates that were relevant in generating the data. The left panel shows results for 100 data sets that includes the disease-related age ({\em diseaseAge}) as a relevant covariate. The center panel shows results for 100 simulations where the disease-related age was not a relevant covariate. The right panel shows distribution of runtimes over the total 200 data sets for both methods. The bar lengths are average runtimes, and the turnstiles indicate runtime standard deviations.} %
\end{figure}

We first confirm that linear mixed modeling cannot capture the covariate relevances whereas our GP modeling approach can, when the covariate effects are nonlinear. We use the \textit{lme4} package \cite{bates2015} for fitting linear mixed effect models, and the \textit{lmerTest} package \cite{kuznetsova2017} for computing $p$-values of the linear model components. The $p$-values are used as the score in ROC analysis. The resulting ROC curves and AUC scores are shown in Fig.~\ref{fig: comparisons}a. It is evident that the linear mixed model approach performs poorly, whereas \textit{lgpr} is consistently more accurate, reaching near-perfect performance when $N=600$.

We also compare our method with the additive Gaussian process model selection method \textit{LonGP} \cite{cheng2019}.
Here we set up a more difficult covariate selection problem with more covariates of different types, and also generate non-stationary disease effects for half of the individuals. Since \textit{LonGP} uses a sequential model search, we cannot compute full ROC curves for it. Therefore we compare performances by counting how often each covariate is selected. \textit{LonGP} tends to select very few covariates, and to have comparable results for \textit{lgpr}, we set a rather low threshold of $T=80$. Fig.~\ref{fig: comparisons}b shows the number of times each method selected different covariates across the 100 simulated data sets for both the case where the disease effect was and was not relevant. We see that \textit{lgpr} can more clearly distinguish the relevant covariates. The total covariate selection accuracy for \textit{lgpr} is $81.4\%$ for data sets where disease-related age is relevant and $87.0\%$ for those where it is not. Corresponding numbers for \textit{lonGP} are $65.0\%$ and $75.2\%$. Furthermore, the average run time per data set is approximately five times smaller for \textit{lgpr} (Fig.~\ref{fig: comparisons}b).

\subsubsection{Heterogeneous and temporally uncertain disease effect modeling} 
\label{s: details_sim_heter}
\begin{figure} % Fig.3
\centering
\includegraphics[width=0.8\textwidth]{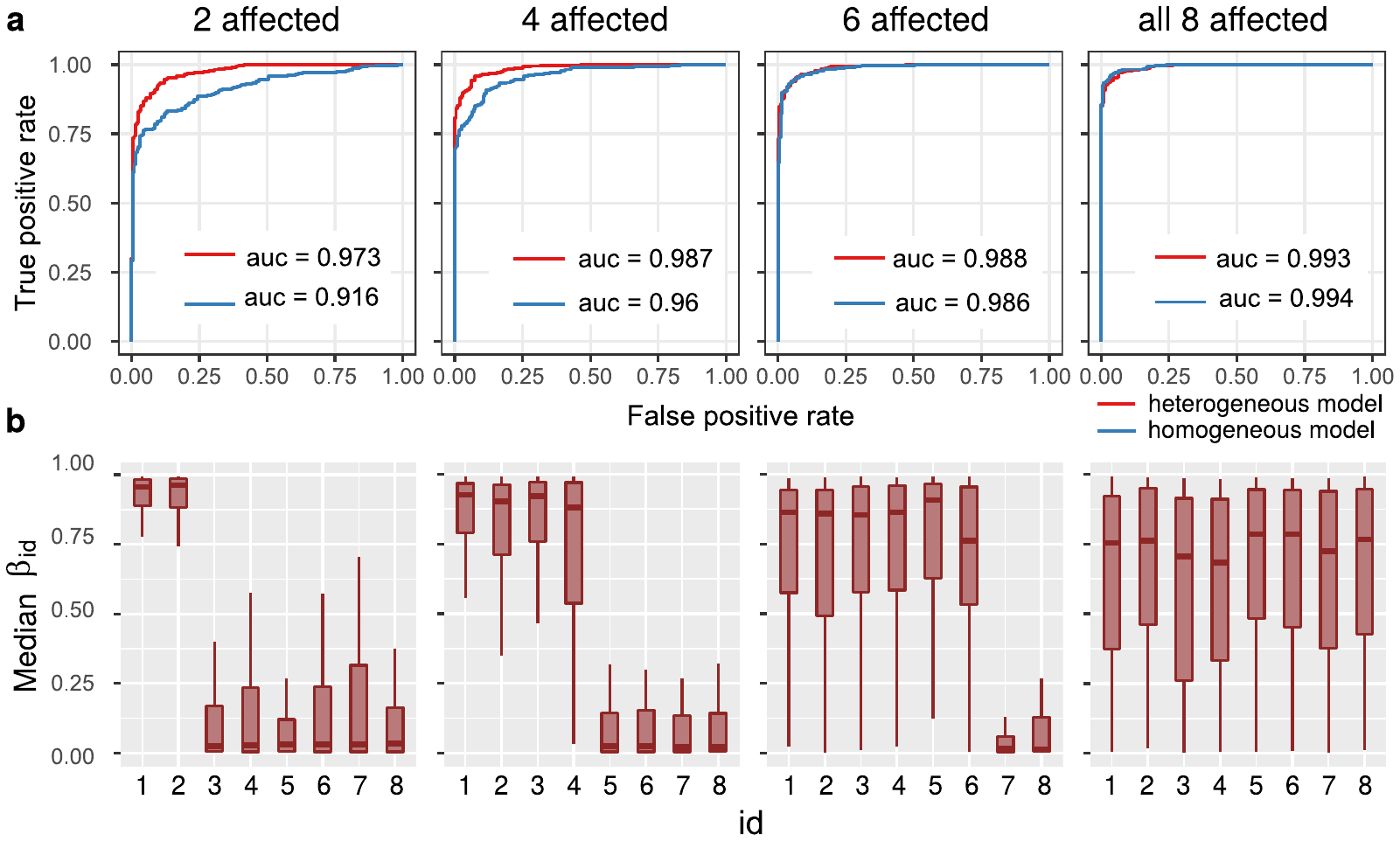}
\caption{\label{fig: roc_heter} Heterogeneous disease effect modeling with \textit{lgpr} improves detection of effects that are present only for a subset of case individuals. \textbf{a)} ROC curves for covariate relevance assessment using both a heterogeneous and a homogeneous disease model for simulated data with 2, 4, 6 and 8 out of the 8 case individuals affected, respectively. \textbf{b)} Heterogeneous modeling with \textit{lgpr} can reveal the affected individuals. The boxplots show the distributions of the posterior medians of the individual-specific disease effect magnitude parameters $\beta_{\text{id}}$, $\text{id}=1, \ldots, 8$ over 100 simulated data sets. The box is the interquantile range ($IQR$) between the 25th and 75th percentiles, vertical line inside the box is the 50th percentile, and the whiskers extend a distance of at most $1.5 \cdot IQR$ from the box boundary.  Each panel corresponds to the same experiment as the one above it.}
\end{figure}

To test the heterogeneous disease effect modeling approach, we generate data with 16 individuals out of which 8 are cases, but so that the disease effect is generated for only $N_{\text{affected}} = $ 2, 4, 6 or 8 of the case individuals. For each data set replication, the inference is done using both a heterogeneous and homogeneous model. The results in Fig.~\ref{fig: roc_heter} show that heterogeneous modeling improves covariate selection accuracy, and the improvement is clearest when $N_{\text{affected}} =  2$. Moreover, in heterogeneous modeling, the posterior distribution of the individual-specific disease effect magnitude parameters $\beta_{\text{id}}$ indicates the affected individuals. See \sfig{4} for a detailed demonstration of heterogeneous model inference.

\begin{figure} %Fig. 4
\centering
\includegraphics[width=0.8\textwidth]{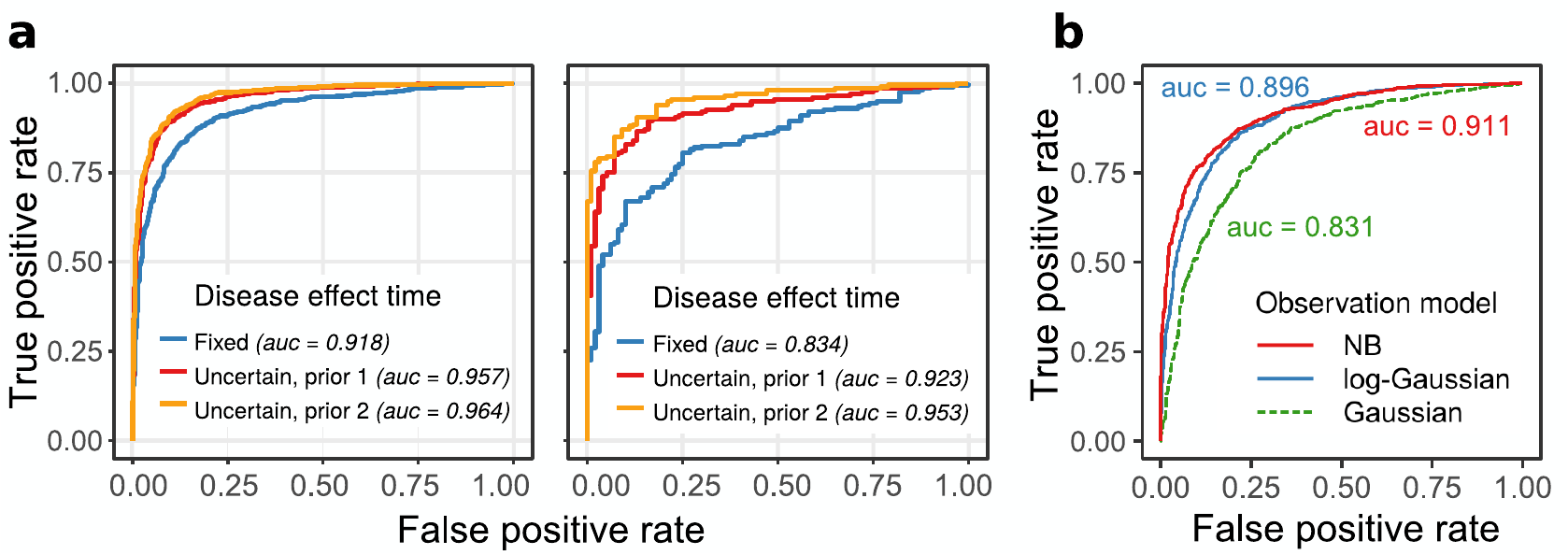}
\caption{\label{fig: roc_nb_et} \textbf{(a)} Modeling the uncertainty in the disease effect time enhances covariate relevance assessment accuracy, when data is generated so that the disease effect can occur earlier than the observed disease initiation. The left panel shows ROC curves for covariate relevance assessment with and without modeling the effect time uncertainty. The red curve is for a model with an exponential decay prior for the difference between the effect time and observed onset \textit{(prior 1)}. The yellow curve is for a model with an oracle prior for the effect time \textit{(prior 2)}. The blue curve is for a model with effect time fixed to equal the observed initiation time. The right panel shows ROC curves for the same three models, in the task of classifying just the disease component as relevant or irrelevant. \textbf{b)} Using a discrete observation model improves covariate selection accuracy for negative binomially distributed count data.}
\end{figure}

To test the model where the disease effect time is considered uncertain, we simulated data where the observed disease initiation time is later than the true generated effect of the disease-related age covariate. For each data set we run the inference first by fixing the effect time to equal the clinically determined onset time, and then using two different priors for the effect time uncertainty. The first prior is $\Delta t \sim \text{Exp}(0.05)$, meaning that the observed onset is most likely, and prior mass decays exponentially towards birth. An oracle prior, which is exactly the distribution that is used to sample the real effect time, is used for reference. The results in  Fig.~\ref{fig: roc_nb_et}a show that the uncertainty modeling improves the covariate selection accuracy, and the oracle prior performs best as expected. Especially, we see that detection of the disease-related age covariate is more accurate when the uncertainty is being modeled. See \sfig{5} for a more specific demonstration of effect time inference.

\subsubsection{Non-Gaussian data}
\label{s: nb_experiment}
To demonstrate the benefit of using a proper observation model for count data, we generate negative binomially distributed data and  run the inference using both a Gaussian and NB observation model. For reference, we also run the inference using the Gaussian observation model after transforming the counts through mapping $y \mapsto \log(1 + y)$. Results in Fig.~\ref{fig: roc_nb_et}b confirm that using the correct observation model in \textit{lgpr} for this kind of count data improves covariate selection accuracy compared to the Gaussian or log-Gaussian models. We note, however, that covariate selection performance of the log-Gaussian model improves (relative to that of the NB model) when data has higher count values and dispersion is smaller, i.e., when the NB model is better approximated by the log-Gaussian model.

\subsection{Longitudinal proteomics data analysis}
\label{s: proteomics}
We used \textit{lgpr} to analyse a longitudinal data set from a recent T1D study \cite{liu2018}, where the longitudinal profile of protein intensities from plasma samples was measured for 11 cases and 10 controls at nine time points that span the initiation of the disease pathogenesis,  %. Protein intensities were measured from plasma samples of 21 children, with nine measurement time points for each child, 
resulting in a total of 189 data points for most proteins. We chose to analyse 1538 proteins which were chosen by requiring that at least 50\% of the measurements must have non-missing values. The exact sample sizes after discarding missing data for each protein are shown in \stab{1}. Eleven children developed T1D, and for those individuals we defined the disease effect time to be the seroconversion age, which was defined as age at the first detection of one or multiple T1D autoantibodies \cite{liu2018}. We performed our modeling using five covariates: id, age, diseaseAge, sex and group (case/control). We followed the preprocessing described in \cite{liu2018} to get normalized protein intensities. Of the categorical covariates, id and sex are modeled as age-dependent category-specific deviations from the shared age effect, and group is a constant group offset variable. 

\begin{figure}% Fig.5
\centering
\includegraphics[width=\textwidth]{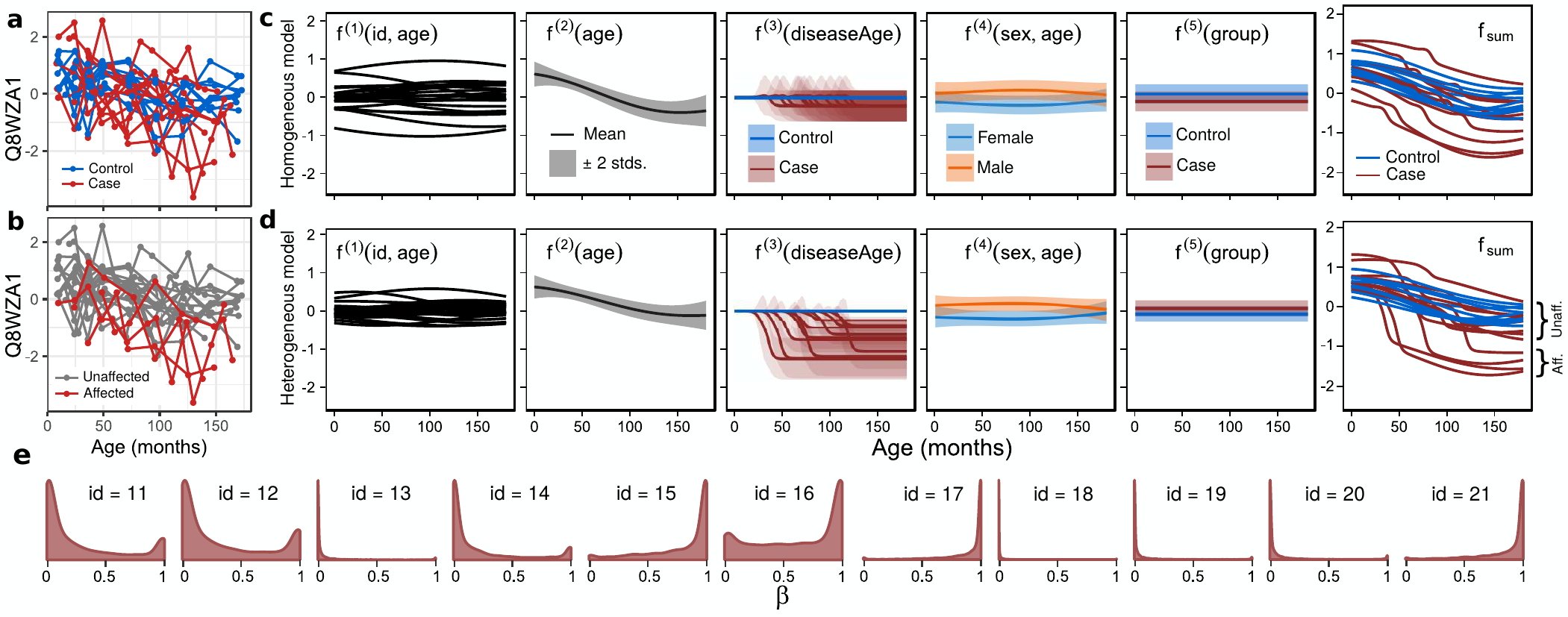}
\caption{\label{fig: main2} Results of analysing one example protein from a longitudinal proteomics data set. \textbf{a)} The normalized measurements for protein Q8WZA1, highlighted based on group (case or control). The lines connect an individual. \textbf{b)} Same data where four case individuals (id=$15,16,17,21$) are highlighted, based on being determined as affected by the disease in heterogeneous modeling. \textbf{c)} Inferred function components, as well as their sum $f$ (using posterior mean parameters), for Q8WZA1 analysed using the homogeneous and \textbf{d)} heterogeneous model. The component relevances (rel$_j$ in Eq.~\ref{eq: rel}) for each $f^{(j)}$, $j = 1, \ldots, 5$ are $0.229, 0.157, 0.03, 0.031, 0.007$ for the homogeneous model and  $0.096, 0.116, 0.25, 0.037, 0.004$ for the heterogeneous model, respectively. The heterogeneous model selects the disease component as relevant, whereas the homogeneous model does not. The posterior distributions of the function components and their sum outside observed time points is computed as explained in \smat. For clarity, standard deviations are not show for $f^{(1)}$ and $f_{\text{sum}}$. \textbf{e)} Kernel density estimates for the posterior distributions of the individual-specific disease effect magnitude parameters of the heterogeneous model.}
\end{figure}

Covariate relevances and selection results for all proteins are included in Tables~S1-S2. As an example, both models confirm the sex association of the Mullerian inhibiting factor (uniprot id P03971) \cite{liu2018}, assigning a relevance score of $0.912$ for the $sex \times age$ interaction term. The homogeneous model finds 38 and the heterogeneous model finds 66 proteins associated with the disease-related age covariate, with intersection of 20 proteins. Fig.~\ref{fig: main2}a shows the normalized measurements for protein Q8WA1 %(\textit{O-linked-mannose beta-1,2-N-acetylglucosaminyltransferase 1}) 
and Figs.~\ref{fig: main2}c-d show the inferred covariate effects using the two different disease effect modeling approaches. The new heterogeneous modeling approach is seen to detect a stronger average disease effect, because it allows the effect sizes to vary between individuals. Moreover, the posterior distributions of individual-specific disease effect magnitude parameters (Fig.~\ref{fig: main2}e), reveal four individuals ($id = 15,16,17,21$) (Fig.~\ref{fig: main2}b), that experience a strong disease effect near the seroconversion time.

\subsection{Longitudinal RNA-seq data analysis}

\label{s: rna-seq}
We analysed also read count data from CD4+ T cells of 14 children measured at $3, 6, 12, 18, 24$ and $36$ months age \cite{kallionpaa2019}. The number of available data points was 6 (for 8 children), 5 (2 children), 4 (2 children) or 3 (2 children), resulting in a total of 72 data points. Seven children had become seropositive for T1D during the measurement interval (cases), while the other seven children were autoantibody negative (controls). We included 519 highly variable genes in our \textit{lgpr} analysis, based on preprocessing steps explained in \smat. We included the same covariates and components in our \textit{lgpr} model as in the proteomics data analysis, and age at the first detection of one or more T1D autoantibodies was again used to compute the disease related age.

\begin{figure} % Fig. 6
\centering
\includegraphics[width=0.85\textwidth]{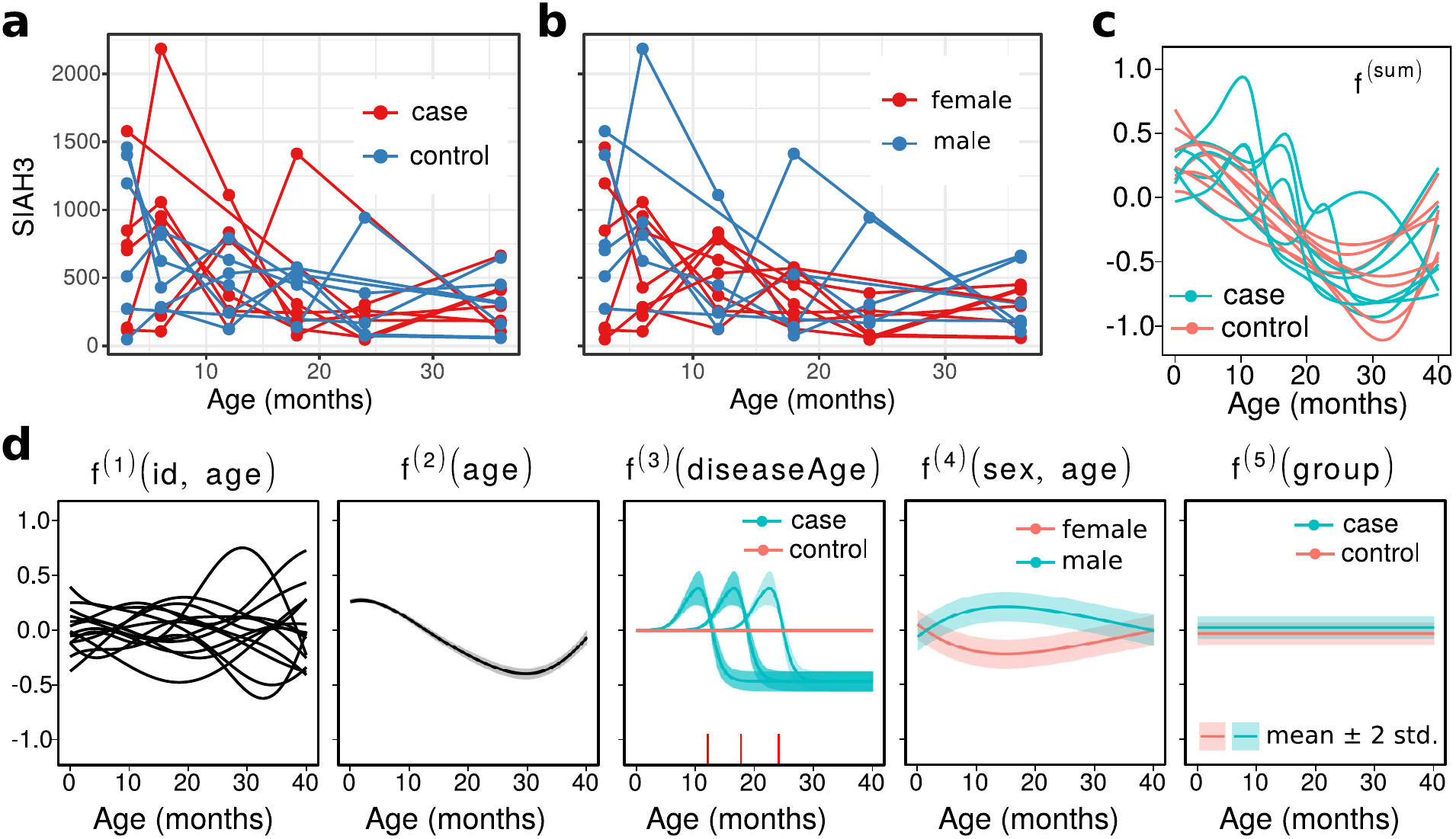}
\caption{\label{fig: gene_425} Data and inferred covariate effects for the SIAH3 gene. \textbf{a)} Raw count data highlighted based on group (case/control) and \textbf{b)} sex. \textbf{c)} Inferred cumulative effect $f^{(\text{sum})}$ and \textbf{d)} additive function components (computed using posterior mean parameters). Interpolation outside observed time points is done as explained in \smat. For clarity, standard deviations are not show for $f^{(1)}$ and $f^{(\text{sum})}$. The seroconversion times of the seven case individuals, i.e. used disease effect times, are 12, 12, 18, 24, 18, 12 and 18 months, indicated by the red vertical lines. Inferred component relevances for $f^{(j)}$, $j = 1, \ldots, 5$ are $0.097, 0.098, 0.077, 0.043, 0.015$, respectively. The selected covariates are are id, age, diseaseAge and sex.}

\end{figure}

Covariate relevances and selection results for all genes are included in \stab{3}. Our analysis confirms the differential expression profile of the IL32 gene between the case and control individuals \cite{kallionpaa2019}, as the group covariate is selected with relevance $0.196$. The disease-related age was initially selected as relevant for a total of 73 genes. As the data is sparse and noisy, we defined a stricter rule and required that the relevance of the disease-related age component alone is larger than $0.05$. This way we detected 12 interesting, potentially disease development-related genes (highlighted in blue in \stab{3}). As an example, Fig.~\ref{fig: gene_425} shows the inferred covariate effects for the SIAH3 (\textit{Seven in absentia homolog 3}) gene.

\section{Conclusions}
\label{s: conclusion}
The \textit{lgpr} tool provides several important novel features for modeling longitudinal data and offers a good balance between flexibility and interpretability. We have shown that the interpretable kernels, heterogeneous disease modeling, uncertainty modeling of effect times, and covariate selection strategy of \textit{lgpr} significantly improve previous longitudinal modeling methods. The tool has an intuitive syntax, and thus provides an easy transition from the standard linear mixed modeling tools to Bayesian non-parametric longitudinal regression. It is widely applicable as the data can involve irregular sampling intervals, different numbers of measurement points over individuals, and crossed categorical factors. Moreover, many types of response variables that are common in postgenomic studies (continuous, discrete, binary, proportion) can be modeled with the proper observation models. The comprehensive software implementation of \textit{lgpr} enjoys state-of-the-art sampling efficiency and diagnostics \cite{vehtari2019} offered by Stan. The user can choose from the numerous presented modeling options and set various parameter priors (which have widely applicable defaults). Overall, \textit{lgpr} has the potential to become a standard tool for statistical analysis of longitudinal data.

\section*{Acknowledgements}
This work has been supported by the Academy of Finland grant no.\ 292660. The authors acknowledge the computational resources provided by the Aalto Science-IT project. 

\section*{Data availability}
No new data were generated or analysed in support of this research.

\bibliographystyle{ieeetr}
\bibliography{ms}

\end{multicols}

\newgeometry{total={7in, 9.25in}}

\clearpage


\section*{Supplementary material}

*lgpr:* An interpretable nonparametric method for inferring covariate effects from longitudinal data - Supplementary material

Juho Timonen, Henrik Mannerström, Aki Vehtari, and Harri Lähdesmäki[*]

Department of Computer Science, Aalto University

October 2, 2020

# Contents

---

[*]Corresponding author.

# 1 Additional related research

Frequentist modeling of longitudinal data has a long history [1, 2]. Random-effect models, marginal models, and conditional models are the three main types of extensions of generalized linear models for longitudinal data, and they make differing assumptions about the joint distribution of the measured responses. There exist also semiparametric and nonparametric approaches, such as local polynomial and spline methods [3]. Generalized linear mixed models [4], which consist of both fixed and random effects, have nevertheless remained very popular. This is because they are an interpretable, fast and powerful model family that usually can be fitted using off-the-shelf software for different types of response variables. The popular R-package *lme4* [5] can be used to fit mixed models using the maximum likelihood or restricted maximum likelihood (REML) method. In practice, however, it is difficult to fit mixed models which involve effects of many crossing categorical covariates, and *lme4* will report a singular fit or imperfect convergence. Hierarchical Bayesian models [6] provide an alternative where the population and group-level parameters have a clearer interpretation, uncertainty can be assessed and posterior diagnostics studied. Their software implementations include *brms* [7, 8] and *rstanarm* [9]. Gaussian process ANOVA [10] is another approach to separate shared effects and interactions of various categorical factors. The *mgcv* package implements generalized additive mixed models [11] based on penalized regression splines.

Despite its importance, the covariate relevance assessment problem for longitudinal data has been largely overlooked. In general, it is often approached by defining a set of alternative models, and recasting it as a model selection problem. Alternatively, significance of different model terms can be tested in a frequentist manner, but for mixed models it is not straightforward to determine the denominator degrees of freedom when computing p-values [12]. Approaches for model selection for linear mixed models, reviewed in [13], include for example information criteria and shrinkage methods. Bayesian models can be compared by information criteria or cross-validation [14], but in general one would need to perform exhaustive search over different models to select the relevant covariates.

The balance between interpretability and flexibility in Gaussian process modeling was initially studied by Plate [15], who proposed to use a sum of univariate and multivariate kernels. An additive kernel that contains interaction terms of all orders was proposed by Duvenaud *et al.* [16] and more complex decomposable additive structures were proposed in [17]. While much of the work on learning GP regression models from data has relied on type-II maximum likelihood estimation, Bayesian techniques have also been proposed in [18, 15, 19]. More recently, semiparametric models that mix linear, spline and GP components [20] as well as more general additive GPs [21, 22, 23] have been developed.

As stated in the main text, additive GP models that are particularly designed for longitudinal study designs include a method of Quintana *et al.* [23] and *LonGP* [24]. *LonGP* uses a stepwise search algorithm for uncovering relevant covariates and was shown to provide accurate covariate selection. However, stepwise search strategies are prone to overfitting at different stages of the search procedure, and the cross-validation based model selection of *LonGP* requires the user to set two different non-trivial threshold values that guide the model search. The stepwise algorithm requires sampling the posterior of GP hyperparameters at each stage, which becomes computationally exhaustive in the presence of many covariates. Furthermore, search algorithms in general will only provide the researcher the information about which model was selected, instead of a relevance measure for all covariates.

Our way of defining the proportion of variance explained by signal is closely related to the recently introduced Bayesian $R^2$-statistic [25]. Like our method, it takes into account the problem

that the classical definition of $R^2$ is not suitable for Bayesian models, since the variance of the predicted values can be larger than the variance of the data. One question that can arise is why not directly use the samples of the kernel hyperparameter $\alpha_j^2$ to divide the proportion of signal between the model components. While they indeed are informative of the effect sizes, we find that studying the posterior of the function components more clearly distinguishes the relevant components. Also the component lengthscales have been used to determine covariate relevance, in an approach that is usually termed automatic relevance determination (ARD) [26]. The motivation is that smaller lengthscales would indicate more relevant components, but ARD has shown to rather measure nonlinearity than relevance [27]. Recently, decomposing the variance explained by different predictors of general machine learning models has also been approached by computing Shapley values, which have game theoretic basis [28].

## 2 Supplementary methods

### 2.1 Observation models

For each data point $i = 1, \ldots, N$, the signal $f$ is transformed though the inverse link function $g^{-1}\left(f(\boldsymbol{x}_i) + c_i\right)$ where $c_i$ are additional scaling values, which can be constant or incorporate normalization factors that are meant to remove technical variation between data points. Biological variation, which we are interested in, is then expressed by the modeled function $f$. The link function $g$ is identity for the Gaussian observation model, logit for binomial and beta-binomial observation models and logarithm for Poisson and negative binomial.

The **Gaussian** observation model for a continuous response is $y_i \sim \mathcal{N}(y_i | f_i, \sigma^2)$, where $f_i = f(\boldsymbol{x}_i)$, and it involves the noise variance parameter $\sigma^2$. In the binomial and beta-binomial observation models for count data, the vector containing the numbers of trials $\boldsymbol{\eta} = [\eta_1, \ldots, \eta_N]^\top$ is supplied as data, along with $\boldsymbol{y}$, which now is the vector of numbers of observed successes. The **binomial** model is $y_i \sim \text{Binomial}(\rho_i, \eta_i)$, where the success probability $\rho_i$ equals to $h_i = \text{logistic}(c_i + f_i)$. In the **beta-binomial** model, $\rho_i$ is random so that

$$\rho_i \sim \text{Beta}\left(h_i \cdot \frac{1-\gamma}{\gamma},\ (1-h_i) \cdot \frac{1-\gamma}{\gamma}\right), \tag{1}$$

and the parameter $\gamma \in [0,1]$ controls overdispersion so that $\gamma \to 0$ corresponds to the binomial model. The **Poisson** observation model for count data is $y_i \sim \text{Poisson}(\lambda_i)$, where the rate $\lambda_i$ equals to $h_i = \exp(f_i + c_i)$. In the **negative binomial** (NB) model, $\lambda_i$ is gamma-distributed with parameters

$$\begin{cases} \text{shape} &= \phi \\ \text{scale} &= \frac{\phi}{h_i} \end{cases}, \tag{2}$$

and $\phi > 0$ controls overdispersion so that $\phi \to \infty$ corresponds to the Poisson model.

When using the Gaussian observation model, we normalize the continuous response measurements $y_i$ to have unit variance and zero mean, and set $c_i = 0$ for all $i$. With binomial and beta-binomial models, we set by default

$$c_i = \text{logit}\left(\frac{1}{N} \sum_{n=1}^{N} \frac{y_n}{\eta_n}\right) \tag{3}$$

and for Poisson and negative binomial models, we set $c_i = \log(\bar{y}) + \log s_i$, where $\bar{y} = \frac{1}{N} \sum_{i=1}^{N} y_i$. Normalization factors $s_i$ are by default 1 for all $i$, but for example TMM values [29] can be used to account for sample normalization.

## 2.2 Gaussian processes

### 2.2.1 Definition

A Gaussian process (GP) is a collection of random variables, any finite number of which has a multivariate normal distribution [30]. A function $f$ is a GP

$$f \sim \mathcal{GP}\left(m(\boldsymbol{x}), k(\boldsymbol{x}, \boldsymbol{x}')\right) \tag{4}$$

with mean function $m(\boldsymbol{x})$ and kernel (or covariance) function $k(\boldsymbol{x}, \boldsymbol{x}')$, if for any finite number of inputs $\{\boldsymbol{x}_p \in \mathcal{X}\}_{p=1}^P$, the vector of function values $\boldsymbol{f} = [f(\boldsymbol{x}_1), \ldots, f(\boldsymbol{x}_P)]^\top$ has a multivariate normal prior $\boldsymbol{f} \sim \mathcal{N}(\boldsymbol{m}, \boldsymbol{K})$ with mean vector $\boldsymbol{m} = [m(\boldsymbol{x_1}), \ldots, m(\boldsymbol{x}_P)]^\top$ and $P \times P$ covariance matrix $\boldsymbol{K}$ with entries $\{\boldsymbol{K}\}_{ik} = k(\boldsymbol{x}_i, \boldsymbol{x}_k)$. The kernel function encodes information about the covariance of function values at different points. Choosing a suitable kernel function therefore is an essential part of GP modeling. New kernels can be created by composing simple base kernels through addition and multiplication.

### 2.2.2 Additive Gaussian processes

We model an additive function $f = f^{(1)} + \ldots + f^{(J)}$ so that each component $j = 1, \ldots, J$ is a GP

$$f^{(j)} \sim \mathcal{GP}\left(0, \alpha_j^2 k_j(\boldsymbol{x}, \boldsymbol{x}')\right), \tag{5}$$

independently from other components. This means that

$$f \sim \mathcal{GP}\left(0, k(\boldsymbol{x}, \boldsymbol{x}')\right) \quad \text{with} \quad k(\boldsymbol{x}, \boldsymbol{x}') = \sum_{j=1}^J \alpha_j^2 k_j(\boldsymbol{x}, \boldsymbol{x}'). \tag{6}$$

We now consider $N$ observations $\{\boldsymbol{x}_n\}_{n=1}^N$. The GP model induces a multivariate normal prior

$$\boldsymbol{f}^{(j)} = \left[f^{(j)}(\boldsymbol{x}_1), \ldots, f^{(j)}(\boldsymbol{x}_N)\right]^\top \sim \mathcal{N}\left(\boldsymbol{0}, \boldsymbol{K}^{(j)}\right) \tag{7}$$

for each vector $\boldsymbol{f}^{(j)}$, $j = 1, \ldots, J$. The matrix $\boldsymbol{K}^{(j)}$ is defined so that its elements are $\{\boldsymbol{K}^{(j)}\}_{ik} = \alpha_j^2 k_j(\boldsymbol{x}_i, \boldsymbol{x}_k)$. This means that the prior for $\boldsymbol{f} = \boldsymbol{f}^{(1)} + \ldots + \boldsymbol{f}^{(J)}$ is $\boldsymbol{f} \sim \mathcal{N}(\boldsymbol{0}, \boldsymbol{K})$, where $\boldsymbol{K} = \sum_{j=1}^J \boldsymbol{K}^{(j)}$.

### 2.2.3 Analytical total and componentwise posteriors

Observations of the response variable are denoted by $\boldsymbol{y} \in \mathbb{R}^N$. Under the Gaussian observation model (see Section 2.1), the posterior of $\boldsymbol{f}$ is also Gaussian $p(\boldsymbol{f} \mid \boldsymbol{y}) = \mathcal{N}(\boldsymbol{f} \mid \boldsymbol{\mu}, \boldsymbol{\Sigma})$, with

$$\begin{cases} \boldsymbol{\mu} &= \boldsymbol{K} \boldsymbol{K}_y^{-1} \boldsymbol{y} \\ \boldsymbol{\Sigma} &= \boldsymbol{K} - \boldsymbol{K} \boldsymbol{K}_y^{-1} \boldsymbol{K}, \end{cases} \tag{8}$$

where $\boldsymbol{K}_y = (\boldsymbol{K} + \sigma^2 \boldsymbol{I})$ [30]. In addition, the posterior distribution of component $j$ is $p\left(\boldsymbol{f}^{(j)} \mid \boldsymbol{y}\right) = \mathcal{N}(\boldsymbol{f}^{(j)} \mid \boldsymbol{\mu}^{(j)}, \boldsymbol{\Sigma}^{(j)})$, where

$$\begin{cases} \boldsymbol{\mu}^{(j)} &= \boldsymbol{K}^{(j)} \boldsymbol{K}_y^{-1} \boldsymbol{y} \\ \boldsymbol{\Sigma}^{(j)} &= \boldsymbol{K}^{(j)} - \boldsymbol{K}^{(j)} \boldsymbol{K}_y^{-1} \boldsymbol{K}^{(j)} \end{cases}. \tag{9}$$

### 2.2.4 Out-of-sample prediction and component visualization

Assuming the Gaussian observation model, the posterior distribution of $\boldsymbol{f}^* = [f(\boldsymbol{x}_1^*), \ldots, f(\boldsymbol{x}_P^*)]^\top$, where $\boldsymbol{x}_p^*$, $p = 1, \ldots, P$ are new test points, is $p(\boldsymbol{f}^* \mid \boldsymbol{y}) = \mathcal{N}(\boldsymbol{\mu}_*, \boldsymbol{\Sigma}_*)$ with

$$\begin{cases} \boldsymbol{\mu}_* &= \boldsymbol{K}_* \boldsymbol{K}_y^{-1} \boldsymbol{y} \\ \boldsymbol{\Sigma}_* &= \boldsymbol{K}_{**} - \boldsymbol{K}_* \boldsymbol{K}_y^{-1} \boldsymbol{K}_*^\top \end{cases}, \tag{10}$$

where $\boldsymbol{K}_*$ is the $P \times N$ kernel matrix between test and data points, and $\boldsymbol{K}_{**}$ is the $P \times P$ kernel matrix between test points. We then obtain the posterior predictive distribution $p(\boldsymbol{y}^* \mid \boldsymbol{y}) = \mathcal{N}(\boldsymbol{\mu}_*, \boldsymbol{\Sigma}_* + \sigma^2 \boldsymbol{I})$ [30]. Also the component posteriors $p\left(\boldsymbol{f}_*^{(j)} \mid \boldsymbol{y}\right)$ are multivariate Gaussian with mean and covariance

$$\begin{cases} \boldsymbol{\mu}_*^{(j)} &= \boldsymbol{K}_*^{(j)} \boldsymbol{K}_y^{-1} \boldsymbol{y} \\ \boldsymbol{\Sigma}_*^{(j)} &= \boldsymbol{K}_{**}^{(j)} - \boldsymbol{K}_* \boldsymbol{K}_y^{-1} \boldsymbol{K}_*^{(j)\top} \end{cases}. \tag{11}$$

Eq. 11 is used for visualizing the covariate effects on a denser time grid than the observed data in the figures of this manuscript. With non-Gaussian observation models, we interpolate the samples of $\boldsymbol{f}^{(j)}$ using kernel regression $\boldsymbol{f}_*^{(j)} = \boldsymbol{K}_*^{(j)} \boldsymbol{K}^{-1} \boldsymbol{f}^{(j)}$.

## 2.3 Probabilistic covariate selection

The selection strategy presented in the main text tackles the task of classifying a component/covariate as relevant or irrelevant. To provide perhaps even more informative and interpretable results, we present here a method for computing a selection probability for each component. The method is based on defining the relevance of component $j$ in sample $s$ as in Eq. 4 of main text and repeating the selection procedure in Eq. 5 of main text for each sample $s = 1, \ldots, S$ separately. The selection probability $p_j(T)$ of component $j$ is then the proportion of times it was selected in the total $S$ selection tasks. This method still requires setting the threshold $T$, but is less sensitive with respect to it than the non-probabilistic selection. In order to avoid having to set a fixed threshold $T$ at all, one can instead compute the selection probabilities as $p_j = \int_0^1 p_j(T) w(T) dT$ where $w(T)$ is a weight distribution that satisfies $\int_0^1 w(T) dT = 1$. Our package provides a routine for this using simple numerical quadrature.

## 2.4 Prior specification

Our software implementation allows the user to set various different priors for the sampled hyperparameters. To allow general prior specifications for the lengthscales, all continuous covariates are scaled to have unit variance and zero mean. An exception to this is the disease-related age, because its lengthscale prior needs to be set only for the warped inputs that lie on the interval $[-1, 1]$. When using the Gaussian observation model, also the response variable is standardized this way for easier specification of priors for the marginal variance and noise variance parameters. The default priors are visualized in Fig. S3. Unless otherwise stated in the experiments, we use a Student-$t_\nu^+$ prior with $\nu = 20$ for all marginal standard deviations, and a Log-Normal$(0, 1)$ prior for the lengthscale parameters. If using the Gaussian observation model, we use an Inverse-Gamma$(2, 1)$ prior for the noise variance parameter. For the input warping steepness, our default prior is Inverse-Gamma$(14, 5)$, which has most of its probability mass between 0.2 and 0.8. This is set based on assuming that the disease-related age is expressed in months. In different applications, prior for the input warping steepness possibly needs to be modified based on the presumed time scale of the disease effect. Fig. S3d-e demonstrate determining a suitable prior for this parameter.

5

## 2.5 Note on computational complexity

An advantage of our methodology is that inferring covariate relevances requires fitting only one model, since the model is interpretable by construction. The computational demand therefore is equivalent to fitting the model in Stan [31] by sampling its posterior. The complexity of this is affected by several factors. First of all, it depends on the number of MCMC iterations needed to explore the posterior distribution sufficiently, which depends on how difficult the global posterior geometry is. Cost of one iteration in the dynamic HMC algorihm of Stan depends on the number of gradient evaluations needed, which is in turn affected by the local posterior geometry. These geometries depend not only on the model but also data. Effort has been made to use parametrizations that are less prone to difficult posterior shapes. The cost of one gradient evaluation is dominated by having to compute the Cholesky decomposition of an $N \times N$ matrix, where $N$ is the number of observations. This in practice limits the scalability to at most a few thousand observations. To facilitate analysis of larger data, a future development direction of *lgpr* is to apply a basis function approximation for the GP covariance function [32].

## 3 Details of simulated data experiments

In each experiment, we generated data with covariates id, age, and different types of continuous and categorical covariates. For each individual, the value of any categorical group covariate was drawn from the binomial distribution with $p = 0.5$. Continuous covariates other than age and disease-related age were generated so that for each subject $i$, the value of covariate $x$ at the $k$th measurement point is $x_{ik} = \sin(r_i \cdot m_k + b_i) + o_i$, where $m_1, \ldots, m_P$ are equispaced on the interval $[0, 2\pi]$ and the values $r_i, b_i, o_i$ were drawn independently for each subject and covariate so that $r_i \sim \text{Uniform}\left[\frac{\pi}{8}, \frac{\pi}{4}\right]$, $b_i \sim \text{Uniform}\left[\pi, 2\pi\right]$ and $o_i \sim \text{Uniform}\left[-0.5, 0.5\right]$. We followed a generative process where we first generated additive components $\boldsymbol{f}^{(j)}$ from GP priors, as illustrated in Fig. 1b of the main manuscript, and scaled them so that the effect size (variance) was one for relevant components and zero for the irrelevant ones. Effects of categorical covariates were always included as deviations from a shared age effect in both simulation and inference. The response variable $\boldsymbol{y}$ was generated from $\boldsymbol{f}$ by adding Gaussian noise, except for in the experiment with non-Gaussian data. We used *lgpr* with four independent MCMC chains, 2000 iterations each, and the first half of each chain was discarded as warmup.

### 3.1 Comparison with linear mixed effect modeling

The 100, 300 and 600 data points were generated so that the number of individuals was 20, 30, and 30, and the measurement times were at $\{12 \cdot m\}_{m=1}^{5}$, $\{6 \cdot m\}_{m=1}^{10}$ and $\{3 \cdot m\}_{m=1}^{20}$ months, respectively. We set the signal-to-noise ratio to only 0.2 so that a lot of data is required to reveal the relevant covariates. We generated data so that it contained an individual-specific and a shared age effect, as well as three categorical covariates $z_1, z_2, z_3$ out of which only $z_1$ was relevant. A lengthscale of 24 months was used to draw the shared age component, and a lengthscale of 12 months for the individual and category-specific age-dependent deviation components. In linear mixed effect modeling, we used the `lmer` model fitting command of the *lme4* package [5] with formula `y ~ 1 + age + (age|id) + z1 + z2 + z3`. Significance of the covariates $z_j$, $j = 1, 2, 3$ was then estimated with the *lmerTest* package [33] by computing the $p$-value of an F-test for single-term deletions with the default Satterthwaite approximation for denominator degrees of freedom. These $p$-values were used as the score in ROC analysis. Since it is not straightforward to compute covariate significances for the id and age covariates using

linear mixed effect modeling, we restricted the covariate relevance assesment problem to only the three categorical covariates $z_j$, $j = 1, 2, 3$.

## 3.2 Comparison with LonGP

In this experiment, we set up a more difficult covariate relevance assessment problem and generated individual-specific, shared, and disease-related age effects. Furthermore, we included four additional categorical covariates $z_i$, $i = 1, \ldots, 4$, that interact with age as category-specific age-dependent deviations from the shared age effect, and four more continuous covariates $x_i$, $i = 1, \ldots, 4$, that have their own shared effect. We generated data with 16 individuals, measured at time points 12, 24, 36, 48, 60 and 72 months, resulting in $N = 96$ data points. In addition, we added Gaussian jitter with $\sigma_t = 1$ to the measurement times to make the data unequally spaced and more realistic. The individual-specific and shared age effects, as well as the effects of $z_1$ and $z_2$ were set to be relevant with lengthscale 24 months for shared and 12 months for other components. Also $x_1$ and $x_2$ were set as relevant with lengthscale 1 (note that these covariates vary on a smaller scale than age, as explained the beginning of this section). The four covariates $z_3$, $z_4$, $x_3$ and $x_4$ were left as irrelevant. Half of the individuals were considered diseased and a disease effect time was drawn for each of them separately, uniformly from the interval $[36, 48]$ months. This time point was used when simulating a nonstationary disease effect, using the kernel in Eq. 4 of the main manuscript, with steepness $a$ drawn from $\mathcal{N}(0.5, 0.1^2)$. However, during inference, we considered the observed effect time to be the next actual measurement time point, as in reality it would be detectable only when visiting a doctor. We first generated 100 data sets where, in addition to the previously described effects, the disease-related age effect is relevant, with lengthscale 1, and then another 100 data sets with no disease effect. Gaussian noise was added so that signal-to-noise ratio is 3.

*LonGP* uses a greedy forward search where at each step, the component that has the best leave-one-out cross-validation or stratified cross-validation score is added to the model if the score exceeds a pre-specified threshold. We used the default thresholds LOOCVR = 0.8 and SCVR = 0.95. By default, the id covariate has to be in the model for *LonGP* and the continuous covariates are searched first before the categorical ones. When categorical covariates are added, also their interaction component with age is added if age is already in the model. If the shared age component is not in the model, the categorical covariates act only as group offsets. In order to create a fair comparison with this procedure, we used *LonGP* so that also the shared age component is by default in the model, and left the id and age covariates out of the selection problem, while their effects were still modeled.

The covariates that were to be classified as either relevant or irrelevant were therefore the categorical covariates $z_i$, $i = 1, \ldots, 4$, continuous covariates $x_i$, $i = 1, \ldots, 4$ and the disease-related age. The hyperparameter priors were same for both models. However, the input warping steepness cannot be inferred in *LonGP*, so we used its default fixed value $a = 0.5$. We used a categorical kernel (returns 1 if categories are same, 0 otherwise) for terms that contain a categorical covariate, because the zero-sum kernel is not an option in *LonGP*.

## 3.3 Heterogeneous modeling of the disease effect

The measurement points in this experiment were 12, 24, 36, 48, 60 and 72 months, and the disease effect time for each diagnosed individual was sampled uniformly from the interval $[46, 48]$ months, but was considered observed at 48 months. In addition to the disease component, we also created relevant individual-specific and shared age effects, and one relevant ($z_1$) and two irrelevant ($z_2$,

7

$z_3$) categorical covariates that interact with age. The used lengthscales were 24 months for shared age, 1 for the disease component and 18 months for other components. All covariates were considered to be subject to selection. We set the signal-to-noise ratio to 3 and simulated a total of 400 data sets, 100 for each case $N_{\text{affected}} = 2, 4, 6, 8$.

### 3.4 Modeling the uncertainty in disease effect time

This experiment involved 12 individuals, 6 of them diseased, measured at $12, 24, 36, 48, 60, 72, 84$ and 96 months and we used a signal-to-noise ratio 1. The components and covariates, as well as the lengthscales were same as in the previous experiment. For each diseased individual $q$, the real effect age $t'_q$ was sampled from the normal distribution with mean 36 and standard deviation 4 months (Fig. S5d). The disease was considered observable at age $\tau_q = \min\{t'_q + t^*_q, 96\}$ months, where $t^*_q$ was drawn from the exponential distribution with rate parameter 0.05 (Fig. S5e). The disease onset was then observed to be at the next measurement time point $t^{\text{obs}}_q = \arg\min_t t \geq \tau_q$, $t \in \{12, 24, 36, 48, 60, 72, 84, 96\}$ and the observed disease-related ages were computed as $x_{\text{disAge}} = x_{\text{age}} - t^{\text{obs}}_q$. We simulated 300 data sets so that the set of true relevant covariates was $\{id, age, z_1\}$ in the first 100, $\{age, diseaseAge, z_1\}$ in the next 100, and $\{id, age, diseaseAge, z_1\}$ in the last 100 data sets.

### 3.5 Non-Gaussian data

Here we generated measurements for 10 individuals at timepoints $6, 12, 18, 24, 30, 36, 42$ and 48 months with jitter $\sigma_t = 0.2$ months. The data contained the covariates id, age and additionally two categorical covariates $z_1, z_2$ and two continuous covariates $x_1, x_2$. We used lengthscale 12 months to simulate the shared age component and 6 months for the group and individual specific age effects. We simulated 300 data sets so that the set of true relevant covariates was $\{age, x_1, z_1\}$ in the first 100, $\{id, age, x_1, z_1\}$ in the next 100, and $\{x_1, x_2\}$ in the last 100 data sets. The process $\boldsymbol{f}$ was generated from the additive GP prior and scaled to have variance 2. Noisy measurements $y_i$ were then drawn from $y_i \sim \text{Poisson}(\lambda_i)$, where the rate $\lambda_i$ was draw from the gamma distribution with shape $= \phi$ and scale $= \frac{\phi}{h_i}$, with $h_i = \exp(-1 + f_i)$ and $\phi = 2$. This created integer-valued data with a large proportion of zeros. During inference using the negative binomial model, we used the default Log-Normal$(1, 1)$ prior of *lgpr* for the dispersion parameter $\phi$.

## 4 Details of real data analysis

### 4.1 Longitudinal proteomics data

For *diseaseAge*, the variance mask kernel in Eq. 4 of the main text was used. We used the Gaussian observation model for the normalized protein intensities, and performed the analysis for the 1538 proteins separately, using both the homogeneous and the heterogeneous disease effect modeling approach (Eq. 5 of the main manuscript). In the latter case the prior for the additional $\beta$ parameters was Beta$(0.2, 0.2)$. We ran four independent MCMC chains, using 4000 iterations for each protein, discarding the first half of each chain as warmup. For a handful of proteins the mixing of chains was slower ($\hat{R}$-statistic [34] $> 1.05$ with initial number of iterations), and we reran the sampling, doubling the number of iterations. We used the 95% threshold ($T = 95$) for covariate selection. In order to avoid fitting to artifacts in the data, we interpreted the heterogeneous modeling results so that the disease-related age covariate was selected only if it

8

was found for at least three case individuals (median of the individual-specific disease effect magnitude parameter $> 0.5$).

## 4.2 Longitudinal RNA-seq data

We used the same steps as in [35] to obtain read counts of genes. We removed genes for which more than 50% of transripts per million (TPM) values were zero, and then filtered out genes for which mean(TPM) $< 3$. After this, the TMM method [29] was used for calculating a normalization factor $s_i$ for each sample, as implemented in the *edgeR* package [36]. The 519 highly variable genes were selected based on highest coefficient of variation for TPM values. We analyzed the count data with the negative binomial observation model, using default priors of *lgpr* for most parameters. As an exception, due to the shorter time span of this data, we used a non-default input warping steepness $a$ prior which allows disease effects that occur on a time interval of approximately 18 months (Fig. S3e). For each gene, we ran four MCMC chains for 3000 iterations. The chains mixed sufficiently well for all genes ($\hat{R} < 1.05$).

## 5 Proof of the zero-sum property

**Theorem 1.** Let $r \in \{1, \ldots, M\} = \mathcal{M}$, where $M \geq 2$. We define a kernel function

$$k_{\text{zerosum}}(r, r' \mid M) = \begin{cases} 1 & \text{if } r = r' \\ -\frac{1}{M-1} & \text{otherwise} \end{cases}. \tag{12}$$

Let $k_{\text{base}} : (\mathcal{X}, \mathcal{X}) \to \mathbb{R}$ be another kernel function. Furthermore, we denote their product kernel by $k = k_{\text{zerosum}} k_{\text{base}}$. Now assume that function $f : (\mathcal{X}, \mathcal{M}) \to \mathbb{R}$ has GP prior $f \sim \mathcal{GP}(0, k)$. Then for any $t \in \mathcal{X}$, the prior distiribution of the sum $g(t) = \sum_{r=1}^{M} f(t, r)$ is the Dirac delta distribution at zero.

**Proof.** We study the product kernel in the limit $k = \lim_{\epsilon \to 0^+} k_\epsilon$, where

$$k_\epsilon(\boldsymbol{x}, \boldsymbol{x}') = k(\boldsymbol{x}, \boldsymbol{x}') + \epsilon \cdot \mathbf{1}(\mathbf{x}, \mathbf{x}'), \tag{13}$$

and $\mathbf{1}(\mathbf{x}, \mathbf{x}')$ is one if the inputs are the same data point, and 0 otherwise. Assume that $f$ has GP prior $f \sim \mathcal{GP}(0, k_\epsilon)$ and $\epsilon > 0$. For any fixed $t \in \mathbb{R}$, we have $\boldsymbol{f}_t \sim \mathcal{N}(0, \boldsymbol{K}_\epsilon)$, where

$$\boldsymbol{f}_t = \begin{bmatrix} f(t, 1) \\ \vdots \\ f(t, M) \end{bmatrix} \quad \text{and} \quad \boldsymbol{K}_\epsilon = \begin{bmatrix} k_{\text{base}}(t,t) + \epsilon & & -\frac{k_{\text{base}}(t,t)}{M-1} \\ & \ddots & \\ -\frac{k_{\text{base}}(t,t)}{M-1} & & k_{\text{base}}(t,t) + \epsilon \end{bmatrix}. \tag{14}$$

Since $g(t) = \sum_{r=1}^{M} f(t, r) = \mathbf{1}^\top \boldsymbol{f}_t$, we get

$$g(t) \sim \mathcal{N}\left(\mu, \sigma_\epsilon^2\right), \tag{15}$$

where $\mu = \mathbf{1}^\top \mathbf{0} = 0$ and $\sigma_\epsilon^2 = \mathbf{1}^\top \boldsymbol{K}_\epsilon \mathbf{1} = M \cdot \epsilon$. Therefore, the limit distribution of $g(t)$ is

$$\lim_{\epsilon \to 0^+} \mathcal{N}\left(0, \sigma_\epsilon^2\right) = \lim_{\epsilon \to 0^+} \mathcal{N}\left(0, M \cdot \epsilon\right) = \delta_0, \tag{16}$$

i.e. the Dirac delta distribution.

**Corollary.** With any data, also the posterior distribution of $g(t)$ is $\delta_0$.

9

# 6 Supplementary Figures S1-S5

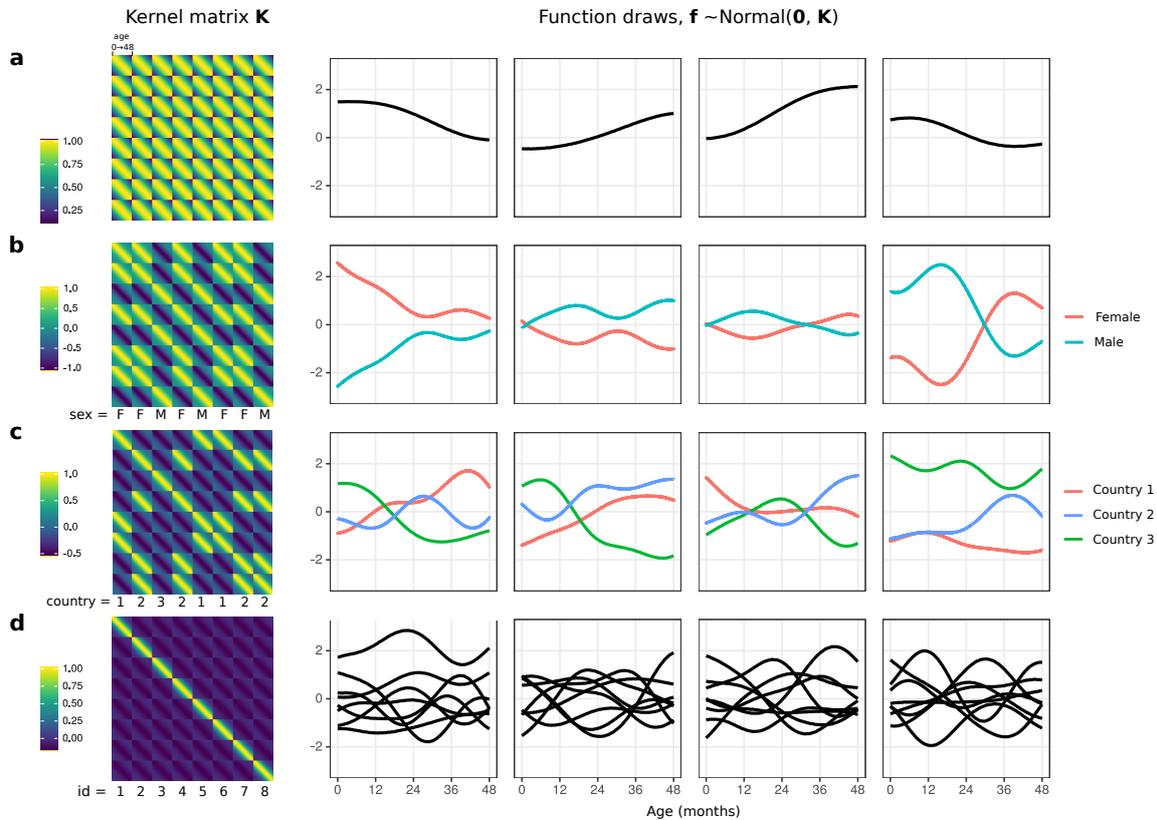

**Fig. S1. Illustration of the zero-sum kernel for modeling category-specific time-dependent deviations from a shared age profile.** Each row shows a kernel matrix and four randomly drawn function realizations for 8 individuals on age span 0 to 48 months. **a)** Standard squared exponential kernel for modeling shared age effects, with lengthscale 24 months. **b)** Interaction of exponentiated quadratic and zero-sum kernel with two categories (Female, Male) and lengthscale 12 months. **c)** Interaction of exponentiated quadratic and zero-sum kernel with three categories (Country 1, Country 2, Country 3) and lengthscale 12 months. **d)** Interaction of exponentiated quadratic and zero-sum kernel with 8 categories (id = 1, . . . , 8) and lengthscale 12 months, for modeling individual-specific time-dependent deviations from the shared age profile.

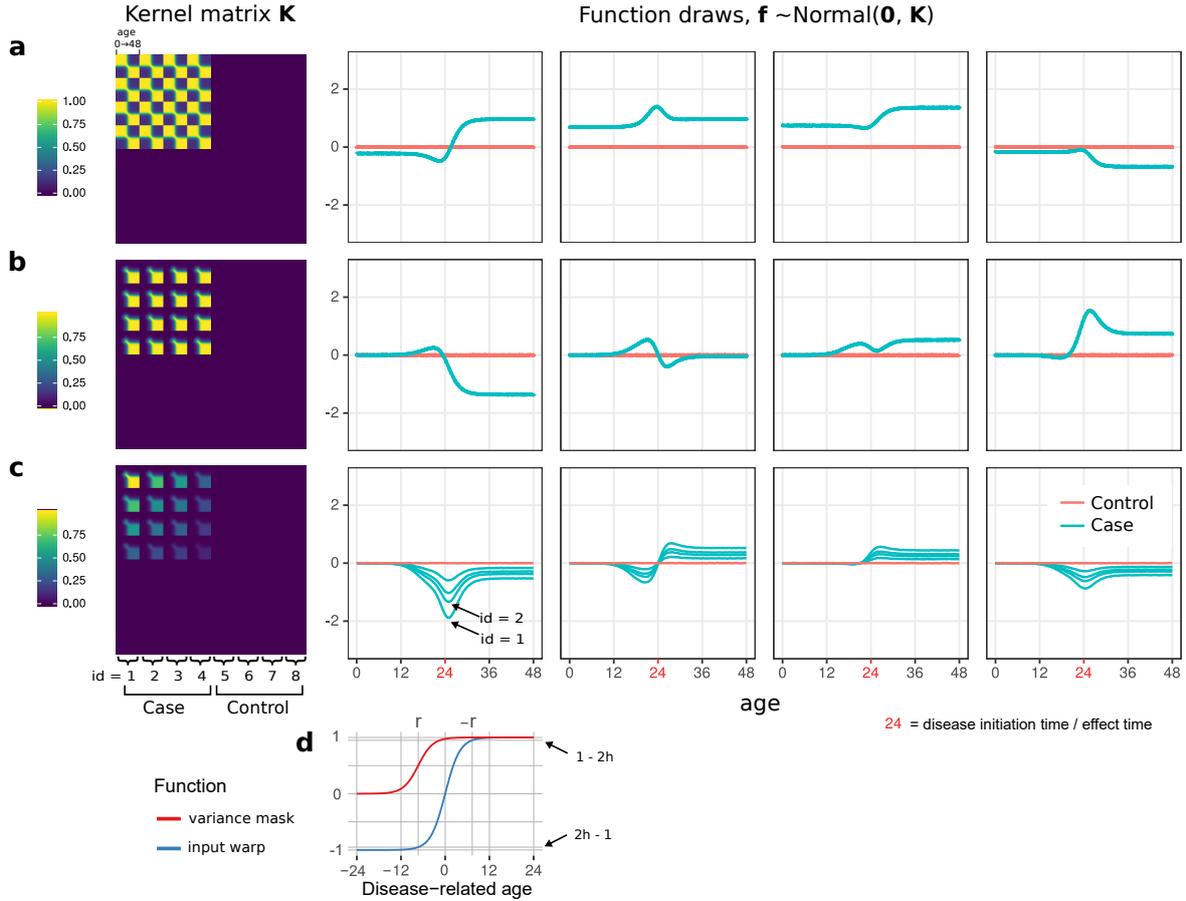

**Fig. S2. Illustration of different kernels for modeling a nonstationary disease effect.** Each row shows a kernel matrix and four randomly drawn function realizations for 8 individuals (4 case, 4 control) on age span 0 to 48 months. Lengthscale $\ell = 1$ and warping steepness $a = 0.5$ have been used for all kernels and the disease effect time is at 24 months. **a)** Standard nonstationary kernel using the input warping approach. The input warping allows the function to vary in time only near the disease effect time for cases, but also allows a baseline difference between cases and controls. **b)** The new variance-masking kernel, which does not allow any difference between groups until close to the effect time and after it. **c)** Heterogeneous version, which in addition allows the magnitude of the disease effect to differ between case individuals. Individual-specific disease effect magnitude parameters have been set to $\beta_1 = 1, \beta_2 = 0.5, \beta_3 = 0.3, \beta_4 = 0.1$. **d)** The input warping and variance masking functions.

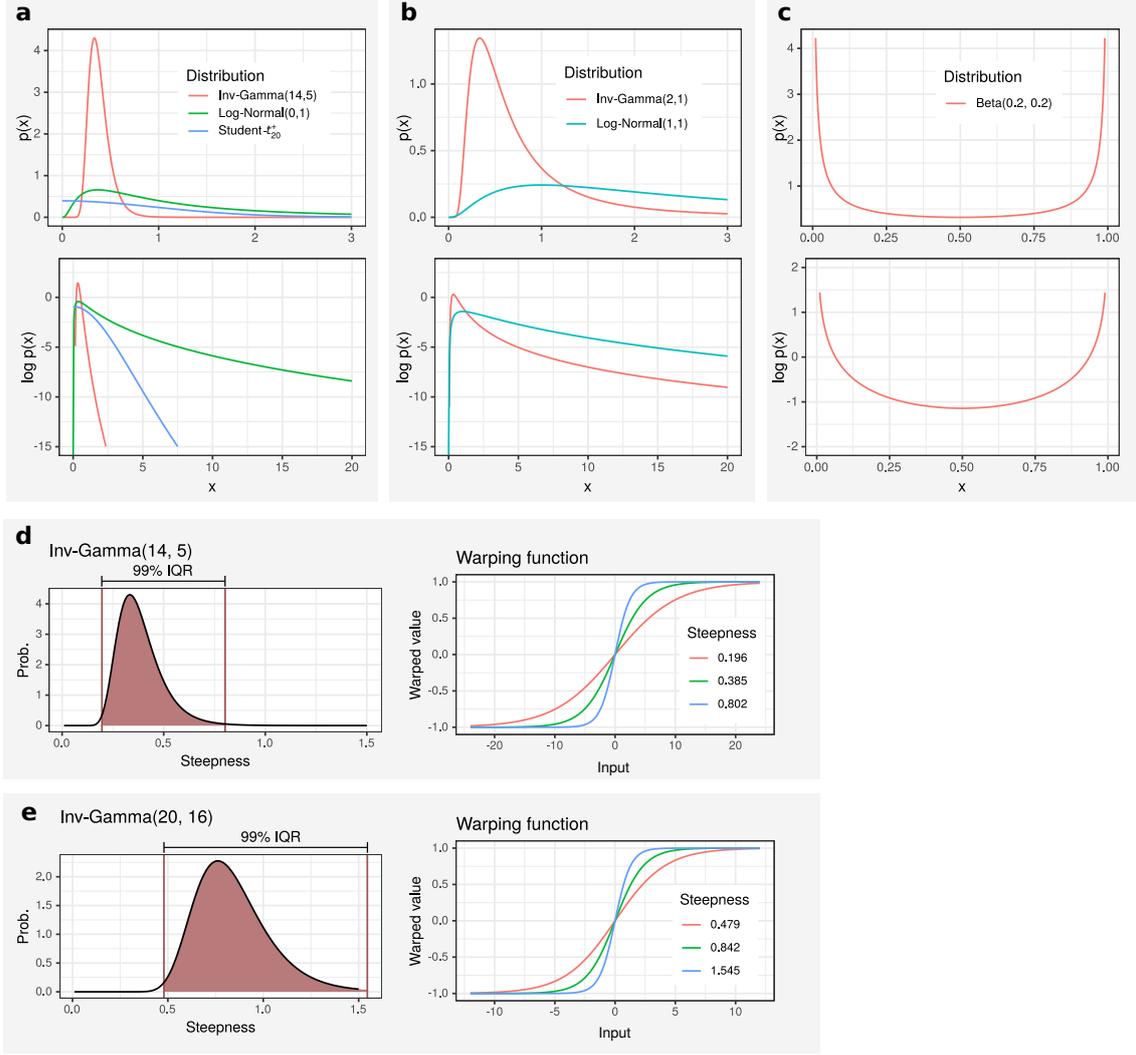

**Fig. S3. Visualization of the prior distributions used for model parameters.** For **a)-c)**, the top row shows probability density functions of the distributions and the second row shows the corresponding densities on log-scale. **a)** Prior distributions for kernel function parameters. A half Student-$t$ prior is used for the marginal standard deviation parameters $\alpha$, because it has a long tail but its mode is at zero. A log-normal prior is suitable for the lengthscales $\ell$ because it does not allow values near zero. An inverse gamma distribution is used for the input warping steepness parameter $a$. **b)** Prior distributions for likelihood function parameters. An inverse gamma distribution is used for the noise variance parameter $\sigma_n^2$ when using a Gaussian likelihood model. A log-normal prior is used for the inverse overdispersion parameter $\phi$ of the negative binomial distribution. **c)** A beta prior for the individual-specific disease effect magnitude parameters is used to prefer parameter values close to either zero or one. Here the beta distribution density is plotted at points $0.01 \leq x \leq 0.99$. **d)** The default Inverse-Gamma(14, 5) prior of the input warping steepness corresponds to input warping functions that allow changes in the disease component approximately ±18 months before and after the disease effect time. Right panel shows the warping function for three steepness values, which correspond the 1st, 50th, and 99th quantiles of the Inverse-Gamma(14, 5) distribution. **e)** Inverse-Gamma(20, 16) prior of the warping steepness allows changes in the disease component approximately ±9 months before and after the disease effect time. Right panel shows the warping function for three steepness values, which correspond the 1st, 50th, and 99th quantiles of the Inverse-Gamma(20, 16) distribution.

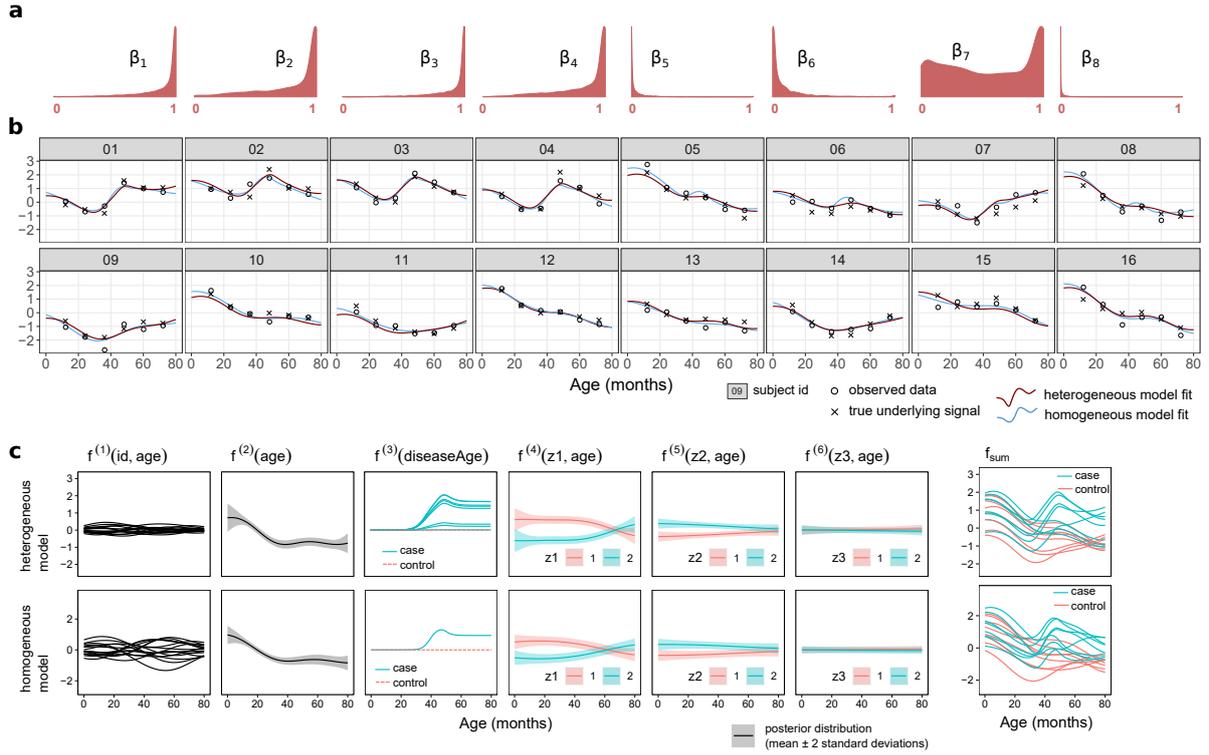

**Fig. S4. Demonstration of inferring a heterogeneous disease effect using simulated data.** Individuals 01-08 are cases, but the disease effect is generated only for individuals 01-04. **a)** Kernel density estimates for the posteriors of the individual-specific disease effect magnitude parameters $\beta_{\text{id}}$, id $= 1, \ldots, 8$. **b)** The data-generating process, noisy data and fits of the heterogeneous and homogeneous models plotted for each of the 16 individuals separately. Model fits are the posterior predictive means, computed using maximum a posteriori (MAP) parameters. **c)** Posterior distributions of each inferred function component for both models, using MAP parameter values. For clarity, standard deviations are not shown for $f^{(1)}$, $f^{(3)}$ and $f_{\text{sum}} = \sum_{j=1}^{6} f^{(j)}$. The true relevant components were $f^{(2)}$, $f^{(3)}$ and $f^{(4)}$. The simulated data of this figure was generated as explained in Section 3.3, except with signal-to-noise ratio 5.

13

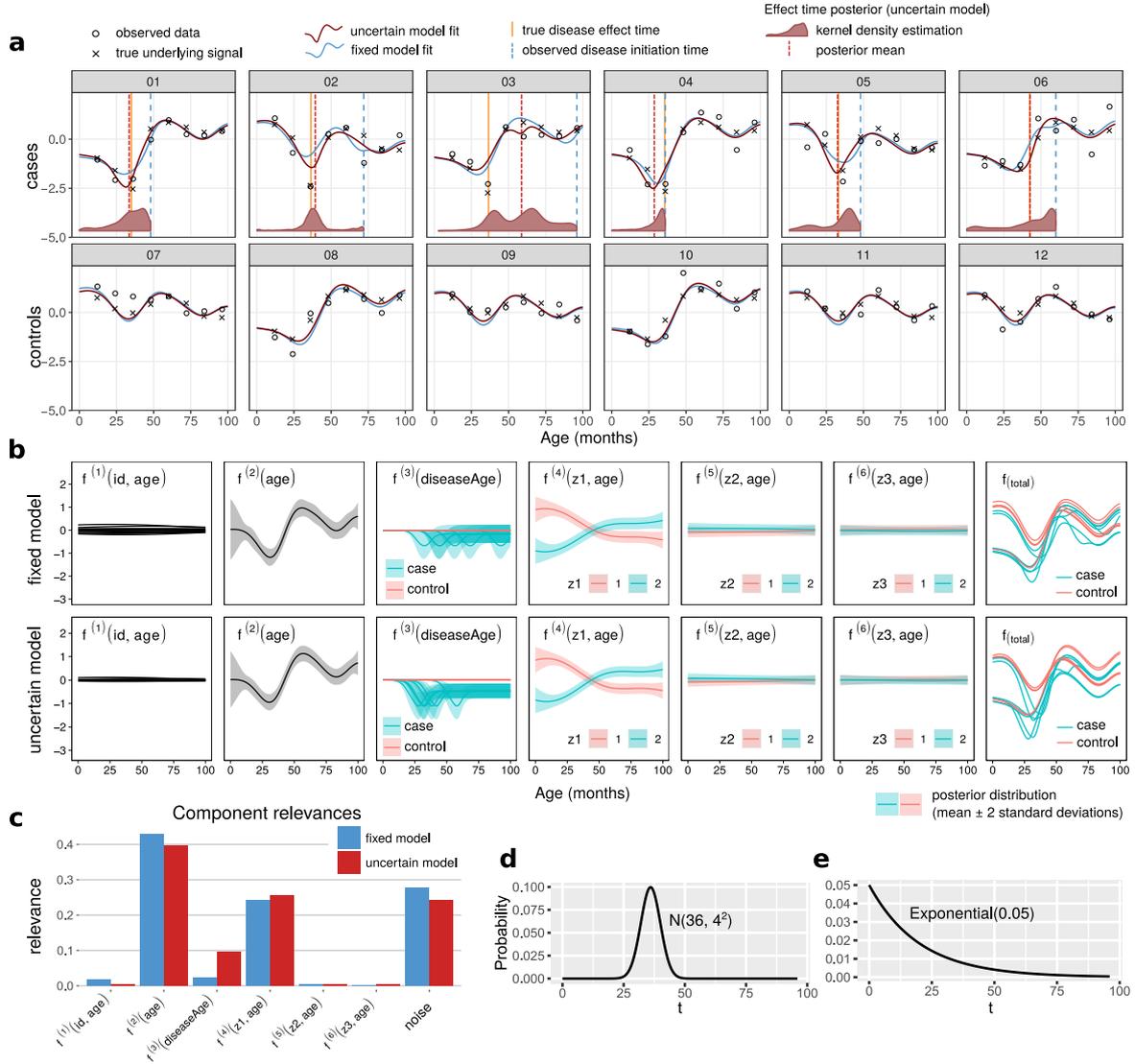

**Fig. S5. Demonstration of inferring the uncertain disease effect time using simulated data.
a)** The data-generating process, noisy data and fits of the uncertain and fixed models plotted for each of the 12 individuals. The fixed model uses the observed disease initiation time (dashed blue line), whereas the uncertain model infers it from data. The uncertain model can more clearly identify the disease effect, because disease initiation is observed later than the true effect occurrence (yellow line). Model fits are the posterior predictive means, computed using posterior mean parameters. **b)** Posterior distributions of each inferred function component for both models, using posterior mean parameter values. For clarity, standard deviations are not shown for $f^{(1)}$ and $f_{(\text{total})} = \sum_{j=1}^{6} f^{(j)}$. **c)** Component relevances inferred by both models for this data realization. The true relevant components were $f^{(2)}$, $f^{(3)}$ and $f^{(4)}$. The simulated data of this figure was generated as explained in Section 3.4, except with signal-to-noise ratio 3. **(d)** The distribution from which the real effect time for each diseased individual is sampled from. **(e)** The distribution from which the disease detection delay is sampled from.

14

[33] A. Kuznetsova, P. B. Brockhoff, and R. H. B. Christensen, "lmerTest package: Tests in linear mixed effects models," *Journal of Statistical Software*, vol. 82, no. 13, pp. 1–26, 2017.

[34] A. Vehtari, A. Gelman, D. Simpson, B. Carpenter, and P.-C. Bürkner, "Rank-normalization, folding, and localization: An improved $\widehat{R}$ for assessing convergence of MCMC," *arXiv:1903.08008*, 2019.

[35] H. Kallionpää, J. Somani, S. Tuomela, U. Ullah, R. de Albuquerque, T. Lönnberg, E. Komsi, H. Siljander, J. Honkanen, T. Härkönen, A. Peet, V. Tillmann, V. Chandra, M. K. Anagandula, G. Frisk, T. Otonkoski, O. Rasool, R. Lund, H. Lähdesmäki, M. Knip, and R. Lahesmaa, "Early detection of peripheral blood cell signature in children developing $\beta$-cell autoimmunity at a young age," *Diabetes*, vol. 68, no. 10, pp. 2024–2034, 2019.

[36] M. D. Robinson, D. J. McCarthy, and G. K. Smyth, "edgeR: a Bioconductor package for differential expression analysis of digital gene expression data," *Bioinformatics*, vol. 26, no. 1, pp. 139–140, 2010.



\end{document}